\documentclass[letterpaper, 10 pt, conference]{ieeeconf}  % Comment this line out if you need a4paper

\IEEEoverridecommandlockouts                              % This command is only needed if 
\usepackage{graphics} % for pdf, bitmapped graphics files
\usepackage{epsfig} % for postscript graphics files
\usepackage{amsmath} % assumes amsmath package installed
\usepackage{amssymb}  % assumes amsmath package installed
\usepackage{color}
\usepackage[dvipsnames]{xcolor}
\usepackage{cite}
\usepackage{booktabs}
\usepackage{multirow}
\usepackage{hyperref}

\renewcommand{\baselinestretch}{0.96}

\newcommand{\methodnamenosp}{MTP}
\newcommand{\methodname}{\methodnamenosp~}
\definecolor{fluorescentorange}{rgb}{1.0, 0.75, 0.0}
\definecolor{canaryyellow}{rgb}{1.0, 0.94, 0.0}
\definecolor{arylideyellow}{rgb}{0.91, 0.84, 0.42}
\definecolor{bananayellow}{rgb}{1.0, 0.88, 0.21}

\title{\LARGE \bf
\methodnamenosp: Multi-hypothesis Tracking and Prediction \\ for Reduced Error Propagation}

\author{Xinshuo Weng, Boris Ivanovic, and Marco Pavone\vspace{-2cm}% <-this % stops a space
\thanks{Xinshuo Weng and Boris Ivanovic are with NVIDIA Research. {\tt\footnotesize \{xweng, bivanovic\}@nvidia.com.} Marco Pavone is with the Department of Aeronautics and
Astronautics, Stanford University, and with NVIDIA Research. {\{\tt\footnotesize pavone@stanford.edu}, {\tt\footnotesize mpavone@nvidia.com}\}.}
}

\begin{document}

\maketitle
\thispagestyle{empty}
\pagestyle{empty}

%%%%%%%%%%%%%%%%%%%%%%%%%%%%%%%%%%%%%%%%%%%%%%%%%%%%%%%%%%%%%%%%%%%%%%%%%%%%%%%%
\begin{abstract}
Recently, there has been tremendous progress in developing each individual module of the standard perception-planning robot autonomy pipeline, including detection, tracking, prediction of other agents' trajectories, and ego-agent trajectory planning. Nevertheless, there has been less attention given to the principled integration of these components, particularly in terms of the characterization and mitigation of cascading errors. This paper addresses the problem of cascading errors by focusing on the coupling between the tracking and prediction modules. First, by using state-of-the-art tracking and prediction tools, we conduct a comprehensive experimental evaluation of how severely errors stemming from tracking can impact prediction performance. On the KITTI and nuScenes datasets, we find that predictions consuming tracked trajectories as inputs (the typical case in practice) can experience a significant (even order of magnitude) drop in performance in comparison to the idealized setting where ground truth past trajectories are used as inputs. To address this issue, we propose a {\em multi-hypothesis} tracking and prediction framework. Rather than relying on a single set of tracking results for prediction, our framework simultaneously reasons about multiple sets of tracking results, thereby increasing the likelihood of including accurate tracking results as inputs to prediction. We show that this framework improves overall prediction performance over the standard single-hypothesis tracking-prediction pipeline by up to 34.2\% on the nuScenes dataset, with even more significant improvements (up to $\sim$70\%) when restricting the evaluation to challenging scenarios involving identity switches and fragments -- all with a relatively minor computation overhead. Our project page is here: \href{https://www.xinshuoweng.com/projects/MTP}{\textcolor{blue}{https://www.xinshuoweng.com/projects/MTP}}.
\end{abstract}

%%%%%%%%%%%%%%%%%%%%%%%%%%%%%%%%%%%%%%%%%%%%%%%%%%%%%%%%%%%%%%%%%%%%%%%%%%%%%%%%
% \vspace{-0.1cm}
\section{INTRODUCTION}

\begin{figure}[t]
\begin{center}
\includegraphics[trim=0cm 9cm 4cm 0cm, clip=true, width=\linewidth]{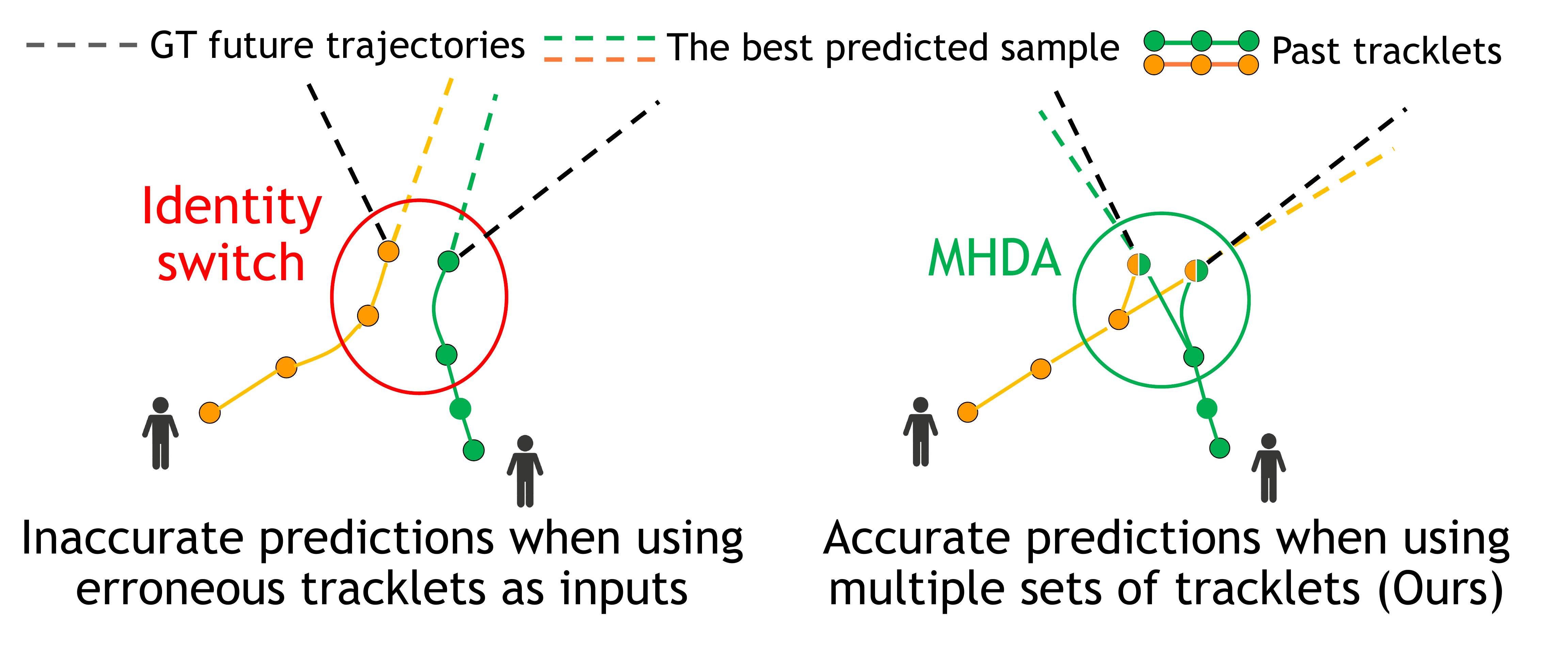}
\vspace{-0.95cm}
\caption{\textbf{(Left)} Using past tracklets as inputs to prediction can cause substantial prediction errors. \textbf{(Right)} By simultaneously reasoning about multiple sets of tracklets via multi-hypothesis data association, one can account for tracking errors and significantly improve prediction performance.}
\label{fig:teaser}
\vspace{-0.5cm}
\end{center}
\end{figure}

% Error propagation in standard tracking and prediction pipeline
Multi-object tracking and trajectory prediction are critical components in modern autonomy stacks. For example, in autonomous driving applications, the outputs of these components are used by the planning module to compute safe and efficient trajectories. Multi-object tracking (MOT) \cite{Zhang2019,Kim2021,Weng2020_AB3DMOT,Wang2020,Benbarka2021,Wang2021_GSDT,Poschmann2020,Guo2020,Weng2020_GNN3DMOT} and prediction \cite{Gupta2018,Kosaraju2019,Eiffert2020,Ivanovic2019,Weng2020_SPF2,Salzmann2020,Yuan2021_AgentFormer,Cao2021,Weng2021_PTP} typically follow a cascaded pipeline, where tracking is performed first to produce past tracklets, followed by a prediction module in charge of predicting other agents' future trajectories. Although such a modularization eases the development cycle, scalability, and interpretability, it also gives rise to significant integration challenges, with cascading errors being a key concern, \emph{e.g.}, a tracking error such as an identity switch can cause a substantial prediction error as shown in Figure \ref{fig:teaser} (left).

% Findings on the error propagation issue
Perhaps surprisingly, the severity of such cascading errors has been relatively under-explored. Indeed, most works on trajectory prediction typically consider the unrealistic setting whereby the prediction module consumes {\em ground truth} (GT) past trajectories as inputs, as opposed to tracklets produced by tracking. In this work, by applying state-of-the-art tracking and prediction methods on the nuScenes \cite{Caesar2019} and KITTI \cite{Geiger2012} datasets, we find that predictions consuming tracklets as inputs experience a {\em significant} performance drop in comparison to the idealized setting where GT past trajectories are used as inputs. Also, if we restrict the evaluation to challenging scenarios involving tracking errors (which are quite frequent, as we will show), prediction errors are increased by up to 28.2$\times$ on KITTI and 17.6$\times$ on nuScenes. The reason for such a significant performance drop is that tracking errors such as identity switch typically induce velocity/orientation estimation errors persisting for a few frames, which can have a detrimental impact on prediction accuracy.

% Our proposed solutions
To address the above issue, we propose a \underline{\textbf{M}}ulti-hypothesis \underline{\textbf{T}}racking and \underline{\textbf{P}}rediction (\methodnamenosp) framework that uses {\em multi-hypothesis} data association to output multiple sets of tracklets as tracking results. Then, these sets of tracklets are used as inputs to the prediction module. The key idea is simple: by simultaneously reasoning about multiple sets of tracklets, the likelihood of including \textit{accurate} tracklets as inputs to prediction is increased (Figure \ref{fig:teaser} (right)). Note that this is different from the standard tracking-prediction pipeline in \cite{Liang2020, Weng2021_PTP,Zhang2019,Kim2021,Weng2020_AB3DMOT,Wang2020,Benbarka2021,Wang2021_GSDT,Poschmann2020,Guo2020,Weng2020_GNN3DMOT}, where only a {\em single} set of tracklets is produced by tracking. In this case, if the past tracklet is off for an object, the prediction might be completely off. 

% Inspiration, experiments, and contribution statement
Our \methodname framework is inspired by the prediction-planning pipeline, where the prediction network typically predicts multiple sets of future trajectories, referred to as trajectory samples in VAE \cite{Ivanovic2019,Salzmann2020,Yuan2021_AgentFormer} or GAN-based \cite{Gupta2018, Kosaraju2019,Eiffert2020} methods. By reasoning about multiple trajectory samples, the likelihood of considering an accurate prediction is higher, thereby enabling a higher level of safety in planning \cite{Rhinehart2021}. \methodname exploits a similar idea, whereby multiple sets of tracklets are generated to improve downstream prediction performance. Through experiments on KITTI and nuScenes, we show that the \methodname framework improves overall prediction performance (up to 34.2\% on the nuScenes dataset), with even more significant improvements (up to $\sim$70\%) when restricting the evaluation to challenging scenarios involving tracking errors. The \methodname framework naturally incurs a computation overhead with respect to its single-hypothesis counterpart, but we show that, fortunately, this overhead is acceptable and still compatible with real-time applications. 

The contributions of this paper are threefold: First, we provide a comprehensive experimental assessment of the impact of tracking errors on prediction performance. Second, we propose the \methodname framework, which aims at reducing error propagation between MOT and prediction by simultaneously reasoning about multiple sets of tracking results. Third, we thoroughly evaluate the performance of \methodname both in terms of prediction accuracy and runtime performance.

% ==================================================================================================================

% \vspace{-0.1cm}
\section{Related Work}\label{sec:rw}

\noindent\textbf{3D Multi-Object Tracking.} Recent online 3D MOT methods often follow a tracking-by-detection pipeline with two steps: (1) Given trajectories associated up to the last frame and detections in the current frame, an affinity matrix is computed, where each entry represents the similarity value between a past trajectory and a current detection; (2) Given the affinity matrix, the Hungarian algorithm \cite{WKuhn1955} is used to obtain a locally optimal matching, which entails making a hard assignment about which past trajectory a current detection is assigned to, so that trajectories can be updated to the current frame. Though significant progress has been made recently for the first step, for example, by improving the affinity matrix estimation using Graph Neural Networks \cite{Chen2020, Weng2020_GNN3DMOT, Li2020} and multi-modal feature learning \cite{Kim2021, Zhang2019}, step two has largely remained the same. In other words, modern 3D MOT methods typically generate a {\em single} set of trajectories via the Hungarian algorithm at inference time, which induces tracking errors that can be detrimental to prediction.

\vspace{1mm}\noindent\textbf{Multi-Hypothesis Data Association.} To improve single-hypothesis MOT, a natural approach is to leverage multi-hypothesis data association (MHDA). The idea is to maintain multiple hypotheses and delay making assignments. As a result, ambiguity in data association can be better resolved in later frames. MHDA has been popular in the 90's and successfully applied to MOT \cite{Kim2015, Long1979, Cox1996} and SLAM \cite{Bernreiter2019, Wang2018}. However, at the time when MHDA was being actively developed, the topic of trajectory prediction was still in its infancy. To the best of our knowledge, our work is the first to adopt MHDA to improve downstream prediction.

\vspace{1mm}\noindent\textbf{Trajectory Prediction.} There has been significant progress on trajectory prediction recently, including \cite{Kitani2012,alahi2016,robicquet2016,Gupta2018,Kosaraju2019,Ivanovic2019,Lee2017,Rhinehart2019,Rhinehart2018}. Yet, almost invariably, such works study the prediction task separately from the 3D MOT task. Specifically, they consider GT past trajectories as inputs to prediction, with no direct accounting of tracking errors. Characterizing and mitigating the propagation of tracking errors to prediction is indeed the key motivation of this paper.

\vspace{1mm}\noindent\textbf{Tracking-Prediction Integration.} A few works have attempted to better couple MOT and prediction tasks. In end-to-end detection and prediction \cite{Liang2020}, the MOT and prediction networks are jointly optimized, which increases performance. Yet, it is still a cascaded, single-hypothesis pipeline, and thus prone to predictions being thrown off by tracking errors. In parallelized tracking and prediction \cite{Weng2021_PTP}, a two-branch tracking and prediction network is proposed. Although this method prevents error propagation in the current frame (tracking results in the current frame are not fed into the prediction branch), it can not do so for the next window of prediction. This is because the method in \cite{Weng2021_PTP} also uses the Hungarian algorithm to generate a single set of tracklets at the current frame, and this can easily lead to tracking errors being propagated to the next window of prediction. In contrast, we replace the Hungarian algorithm with MHDA in the tracking assignment phase, thereby preventing a hard assignment from removing plausible alternative hypotheses. We will show that this idea is quite effective.  

Finally, a concurrent and unpublished work \cite{Yu2021} has also recognized the importance of understanding how tracking errors can impact prediction. Our paper provides a comprehensive quantitative analysis that corroborates the qualitative findings in \cite{Yu2021}, and we propose to leverage MHDA to more robustly accounting for tracking errors, while the solution method in \cite{Yu2021} is still single-hypothesis-based.

\begin{figure}[t]
\begin{center}
\includegraphics[trim=0cm 14cm 14.2cm 0cm, clip=true, width=\linewidth]{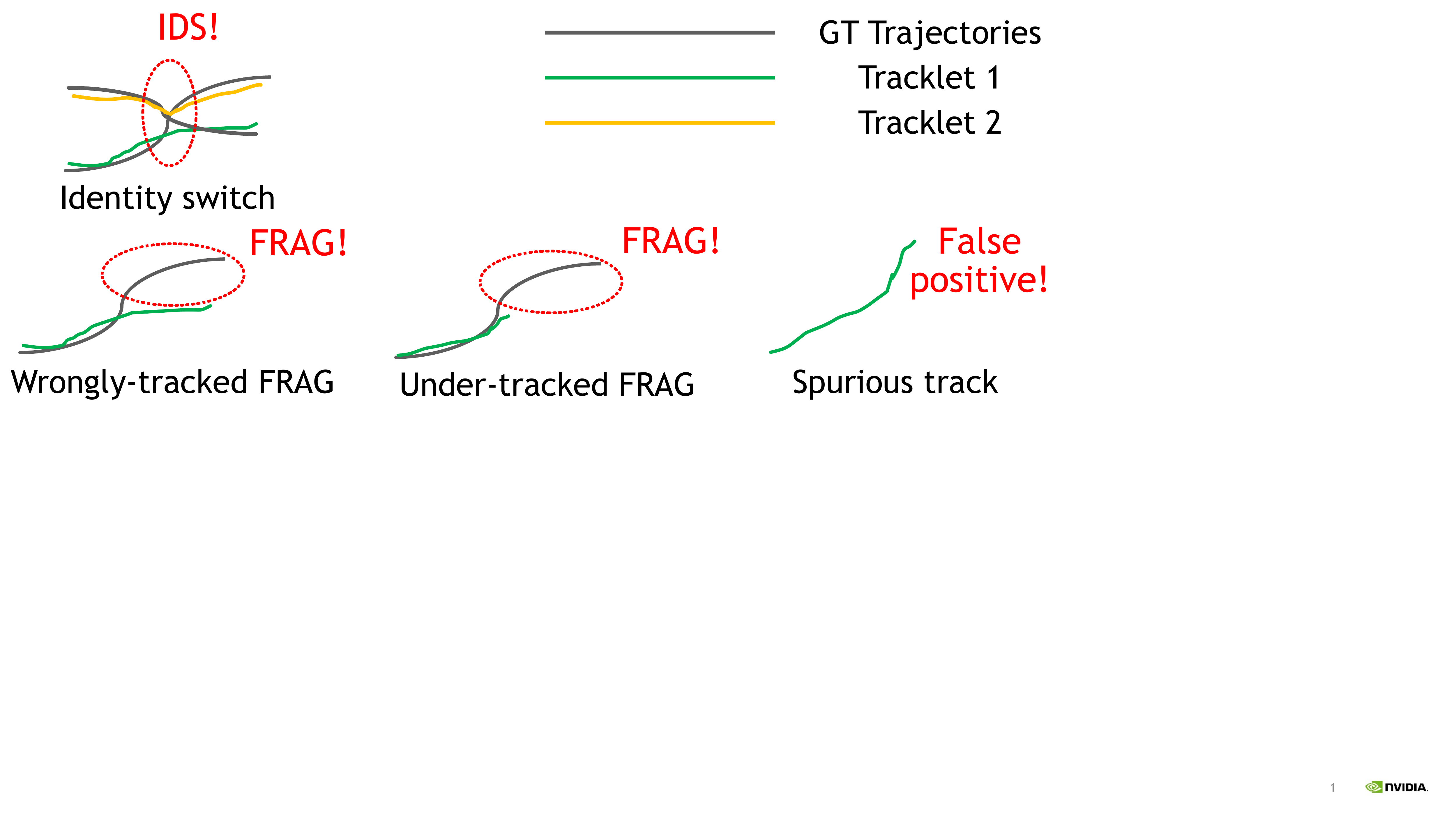}\\
\vspace{-0.5cm}
\caption{The most typical tracking errors are: identity switches \textbf{(top left)}, wrongly-tracked fragments \textbf{(bottom left)}, under-tracked fragments  \textbf{(center)}, and spurious tracks \textbf{(right)}.}     
\label{fig:errors}
\vspace{-0.5cm}
\end{center}
\end{figure}

% \vspace{-0.1cm}
\section{How Do Tracking Errors Affect Prediction?}
% \vspace{-0.1cm}

In this section,  we experimentally study to what extent tracking errors can impact prediction performance. We start by reviewing typical tracking errors. We then outline our methodology, present qualitative and quantitative results, and finally characterize how frequently such errors can arise.

\vspace{-0.1cm}
\subsection{Three key types of tracking errors\label{sec:errors}}

\vspace{1mm}\noindent\textbf{Identity Switches (IDS)} happen if a GT trajectory is matched with two or more different tracklets. For example, as shown in Figure \ref{fig:errors} (top left), the two black GT trajectories are erroneously matched with half of the green and half of the orange tracklets, with a switch in the middle. Such an IDS can happen when two GT trajectories are very close and/or cross each other. IDS can cause large prediction errors as they induce large linear/angular velocity estimation errors, usually persisting for a few frames after the IDS event.

\vspace{1mm}\noindent\textbf{Fragments (FRAG)} refer to GT trajectories that do not closely match any tracklet, either because of a wrong association (wrongly-tracked FRAG, Figure \ref{fig:errors} (bottom left)), or because the detector misses the detection of the object in later frames (under-tracked FRAG, Figure \ref{fig:errors} (bottom center)). 

\vspace{1mm}\noindent\textbf{Spurious Tracks} are tracklets being completely false positives, that is not corresponding to any GT trajectory (Figure \ref{fig:errors} (bottom right)). Spurious tracklets do not affect the recall of the predictions, but do lower the precision. 

\vspace{-0.1cm}
\subsection{Assessment methodology \label{sec:eval}}

In our evaluation, we apply state-of-the-art methods for 3D MOT and prediction, namely AB3DMOT \cite{Weng2020_AB3DMOT} for MOT and PTP for prediction \cite{Weng2021_PTP}\footnote{We use the prediction branch of PTP \cite{Weng2021_PTP}, while obtaining tracking results from \cite{Weng2020_AB3DMOT}, to replicate the standard tracking-prediction pipeline.},  on two standard autonomous driving datasets: KITTI \cite{Geiger2012} and nuScenes \cite{Caesar2019}. 

\vspace{1mm}\noindent\textbf{KITTI.} We use the tracking validation set (see \cite{Scheidegger2018} for the split). We predict 10 future frames using 10 past frames, \emph{i.e.}, 1 second with an FPS of 10. We consider three main object classes labeled in KITTI, \emph{i.e.}, cars, pedestrians, and cyclists. 

\vspace{1mm}\noindent\textbf{nuScenes.} We follow the standard nuScenes prediction challenge guidelines \cite{nuscene_challenge}. Specifically, we use the prediction test set (see nuScenes code \cite{nuscene_code} for the split) for evaluation. We consider vehicle classes, namely car, truck, van, trailer, bus, and construction vehicle. We predict 12 future frames using 4 past frames, \emph{i.e.}, the past 2 seconds with an FPS of 2. 

\vspace{1mm}\noindent\textbf{Evaluation.} We use the standard best of $k$ Average Displacement Error (ADE) and best of $k$ Final Displacement Error (FDE) \cite{Gupta2018} to evaluate prediction\footnote{The ADE is defined as the mean $l_2$ distance between the predicted and GT trajectory; the FDE is defined as the $l_2$ distance between the predicted final position and GT final position at the end of the prediction horizon.}, referred to as minADE$_k$ and minFDE$_k$. While using $k=20$ is standard on KITTI in prior work \cite{Weng2021_PTP}, it is common to use a smaller value of $k$ on nuScenes such as 5 or 10 (see nuScenes leaderboard \cite{nuscene_challenge}). Thus, we use $k=10$ on nuScenes and $k=20$ on KITTI.

To quantify the impact of tracking errors on prediction, a tracking evaluation is needed to find which objects at which frames experience IDS/FRAG errors. To that end, we use the standard 3D tracking evaluation code released in \cite{Weng2020_AB3DMOT}, which aims at matching tracked objects with GT at every frame to see: (1) if there is an identity change of the GT the tracked object is matched to (IDS), (2) if there is a GT that is not matched with any tracked objects (FRAG), and (3) if there is a tracklet not matched with any GT within a threshold (spurious tracks). For the matching threshold, we use a 3D Intersection over Union (IoU)\footnote{Given two boxes, the IoU is defined as the area of their intersection divided by the area of their union. The IoU, by definition, ranges from 0 to 1 and measures how similar the two boxes are. For example, when two boxes exactly overlap with each other, the IoU equals to 1.} of 0.5 on KITTI and a 2D center distance of 2 meters on nuScenes, both of which are standard choices \cite{Weng2020_AB3DMOT, Caesar2019}. 

\definecolor{citrine}{rgb}{0.89, 0.82, 0.04}
\definecolor{goldenyellow}{rgb}{1.0, 0.87, 0.0}
\definecolor{mikadoyellow}{rgb}{1.0, 0.77, 0.05}

\begin{figure}[t]
\begin{center}
\includegraphics[trim=20cm 25cm 21cm 8cm, clip=true, width=0.49\linewidth]{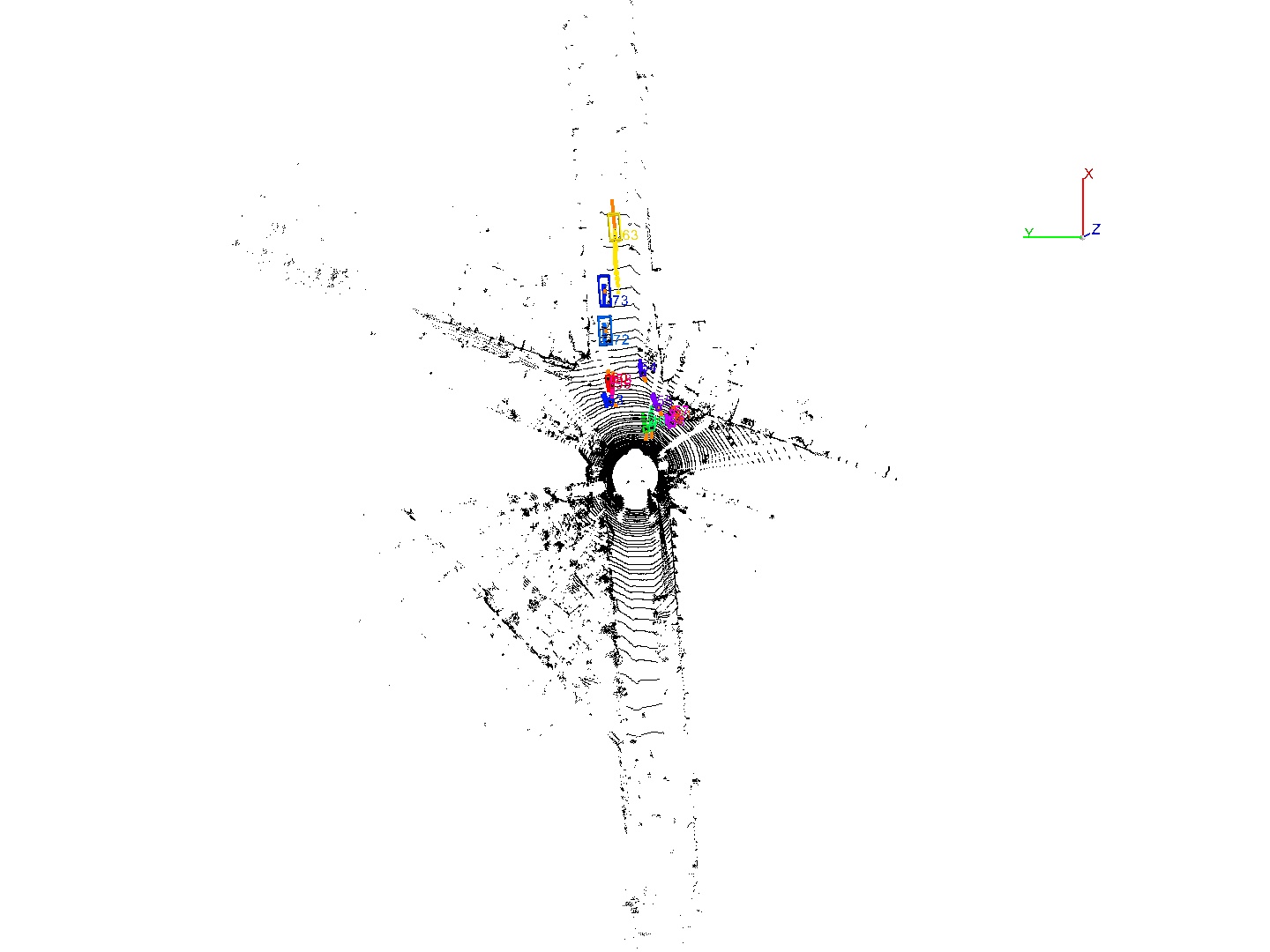}
\includegraphics[trim=20cm 25cm 21cm 8cm, clip=true, width=0.49\linewidth]{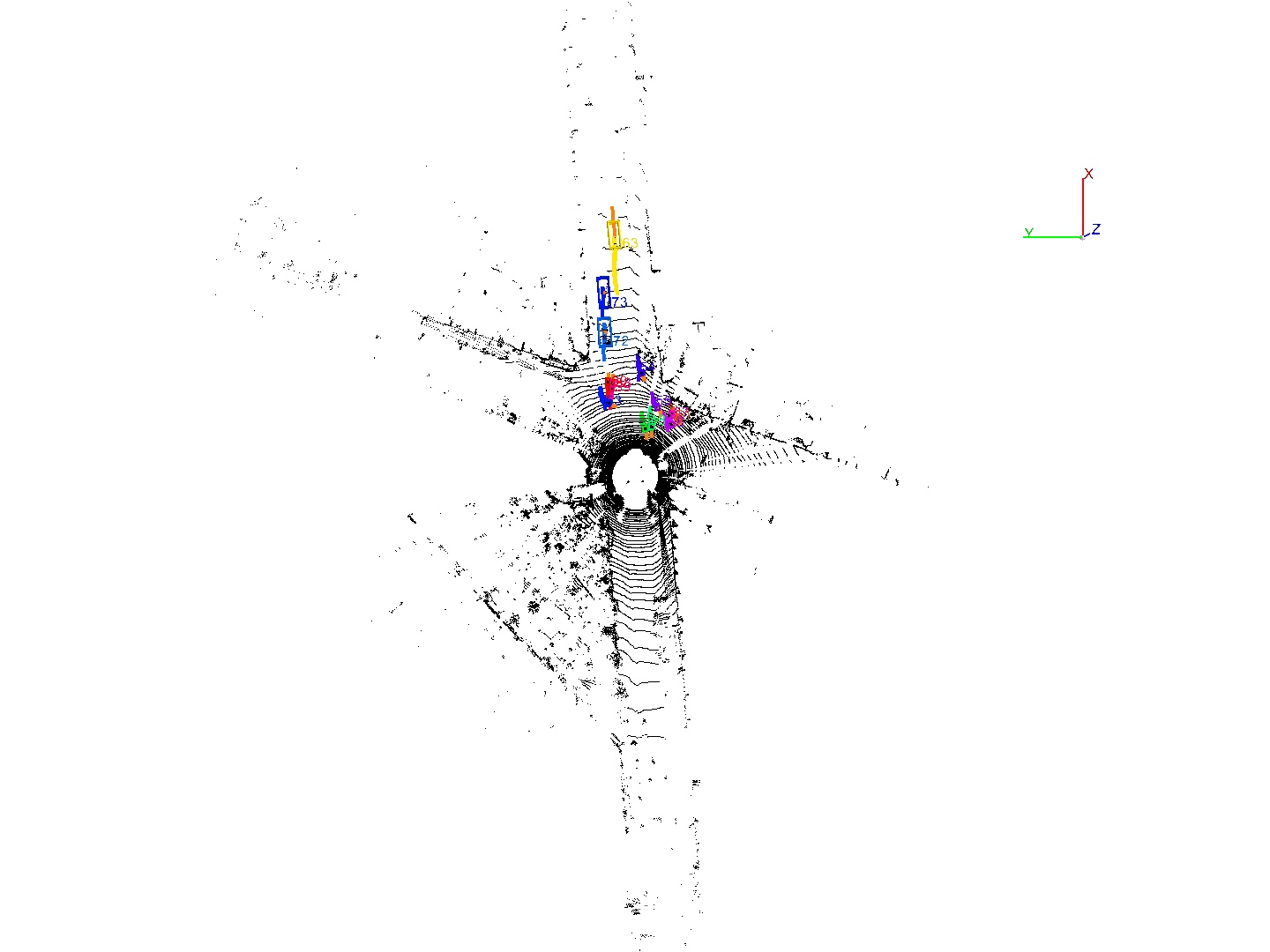}\\
\includegraphics[trim=20cm 25cm 21cm 7cm, clip=true, width=0.49\linewidth]{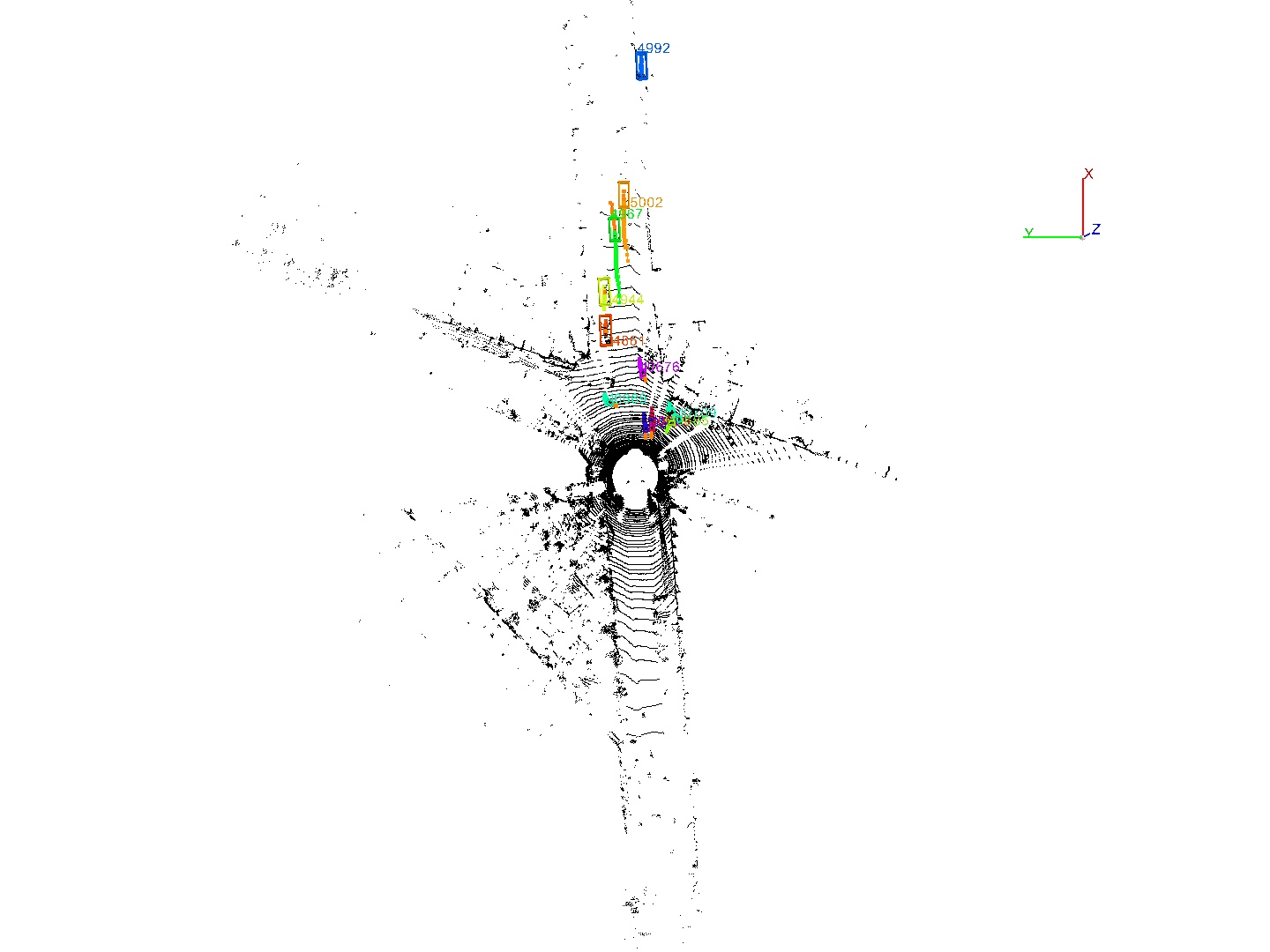}
\includegraphics[trim=20cm 25cm 21cm 7cm, clip=true, width=0.49\linewidth]{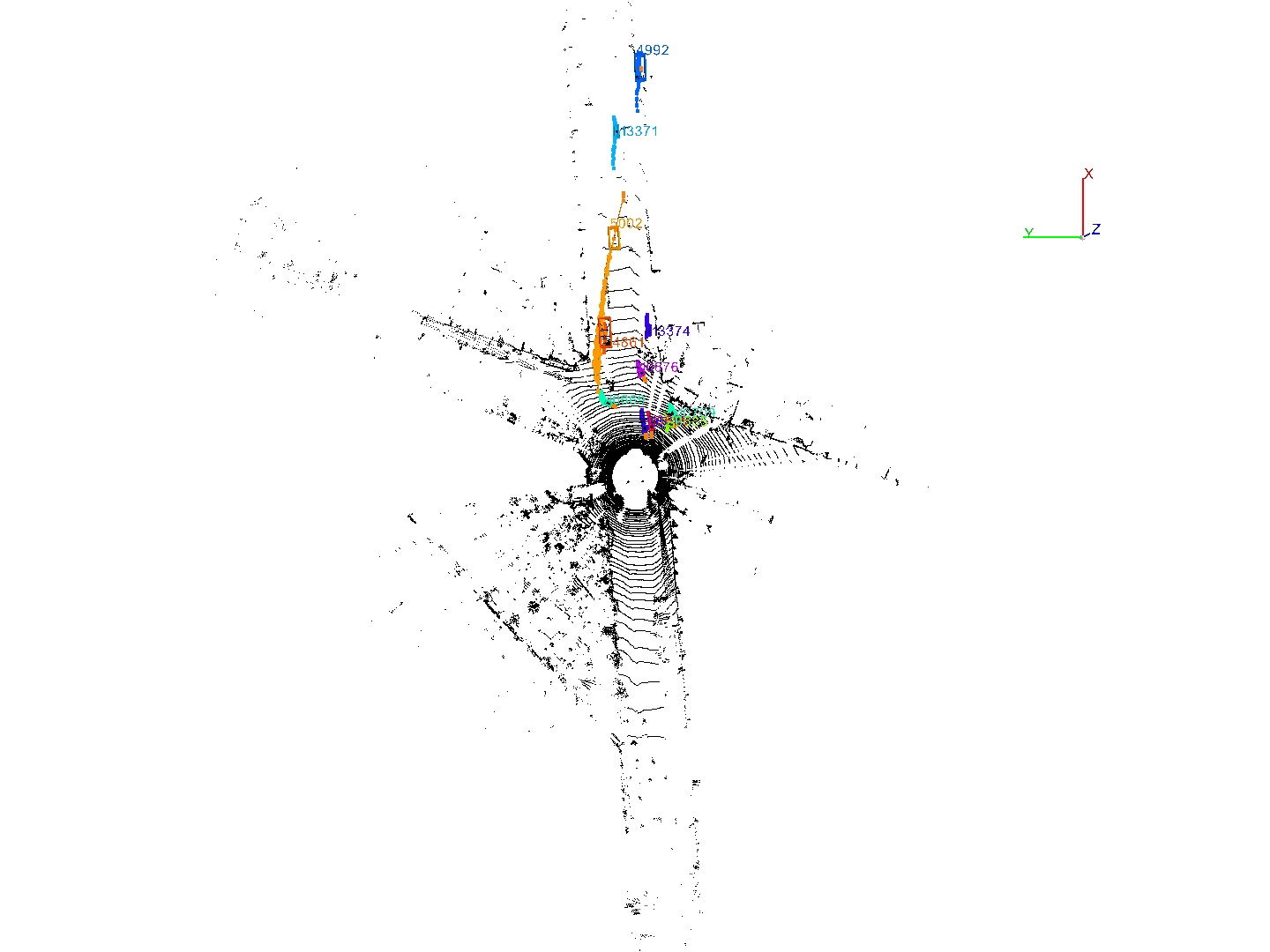}\\
\vspace{-0.35cm}
\caption{IDS causes large prediction errors. \textbf{(Top)}: We show predictions on two frames when using GT past trajectories as inputs. Predictions for the \textcolor{citrine}{yellow} object (ID 63) are accurate in both frames. \textbf{(Bottom)}: We show predictions on the same two frames, but now using past tracklets as inputs. Due to an IDS error, the object ID is switched from \textcolor{Green}{4967} to \textcolor{orange}{5002}, resulting in a velocity estimation error that thwarts predictions.}
\label{fig:IDS_prediction}
\vspace{-0.65cm}
\end{center}
\end{figure}

\begin{figure}[t]
\begin{center}
\includegraphics[trim=10cm 20cm 24cm 10cm, clip=true, width=0.49\linewidth]{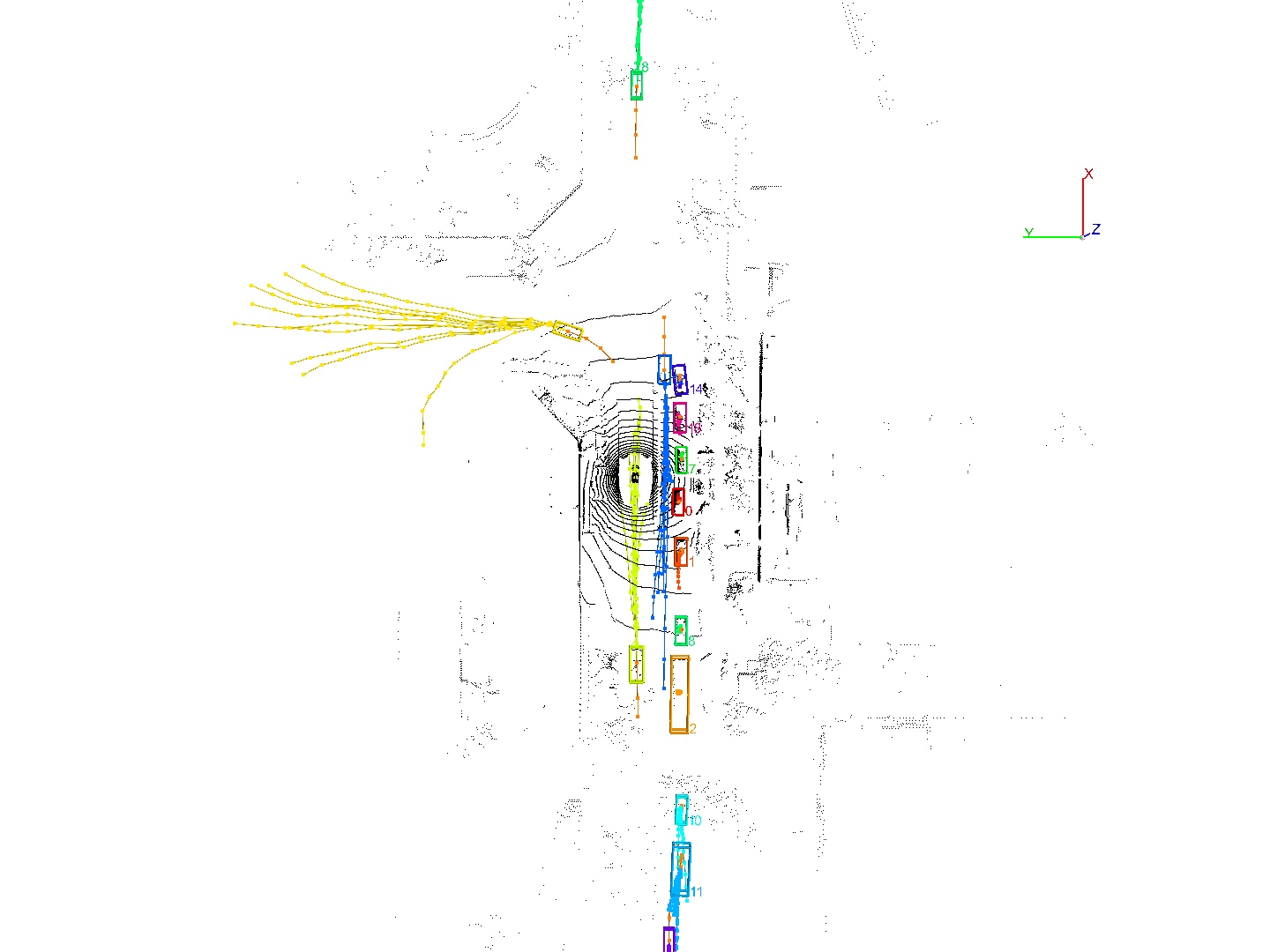}
\includegraphics[trim=10cm 20cm 24cm 10cm, clip=true, width=0.49\linewidth]{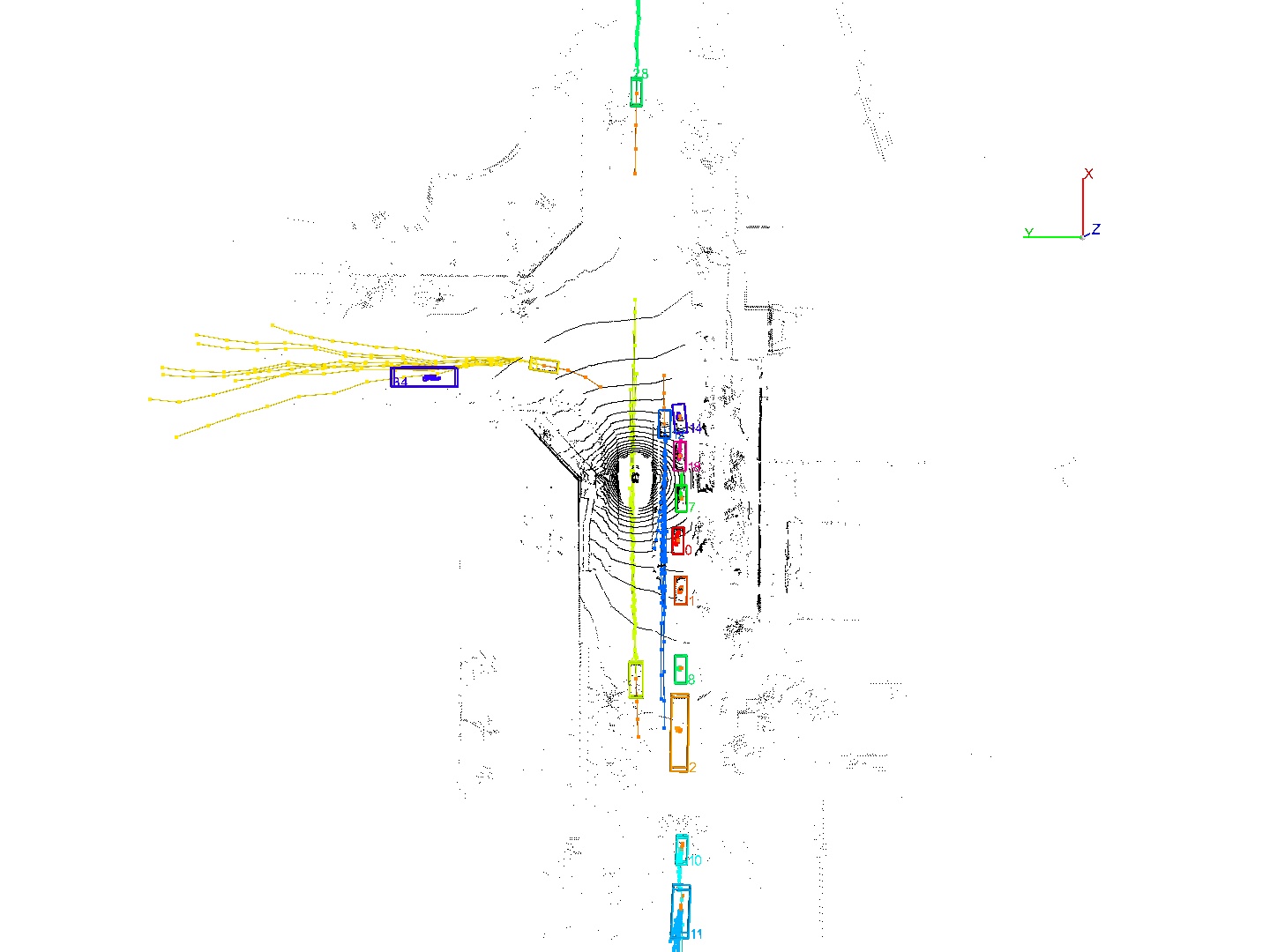}
\includegraphics[trim=10cm 20cm 24cm 10cm, clip=true, width=0.49\linewidth]{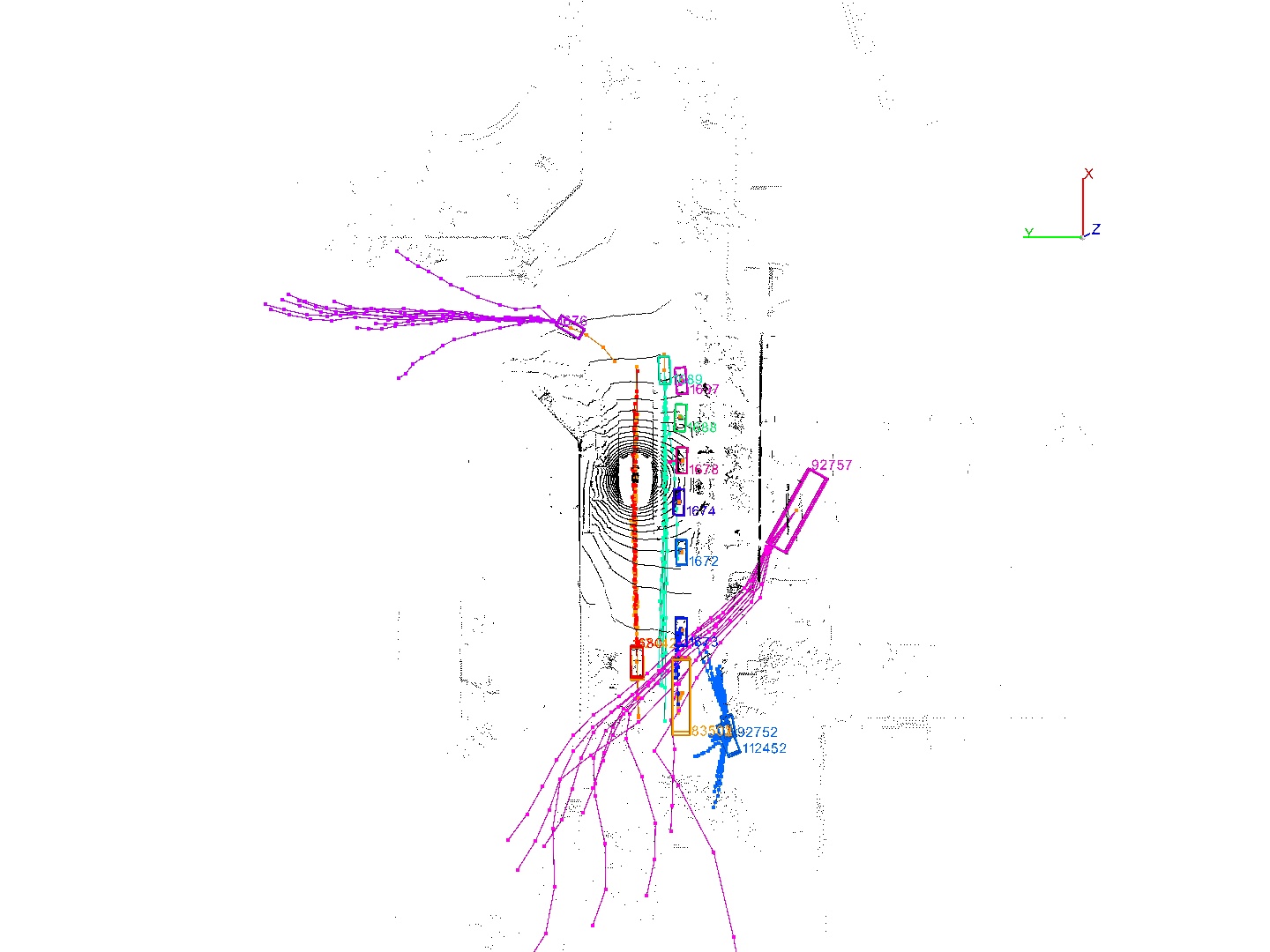}
\includegraphics[trim=10cm 20cm 24cm 10cm, clip=true, width=0.49\linewidth]{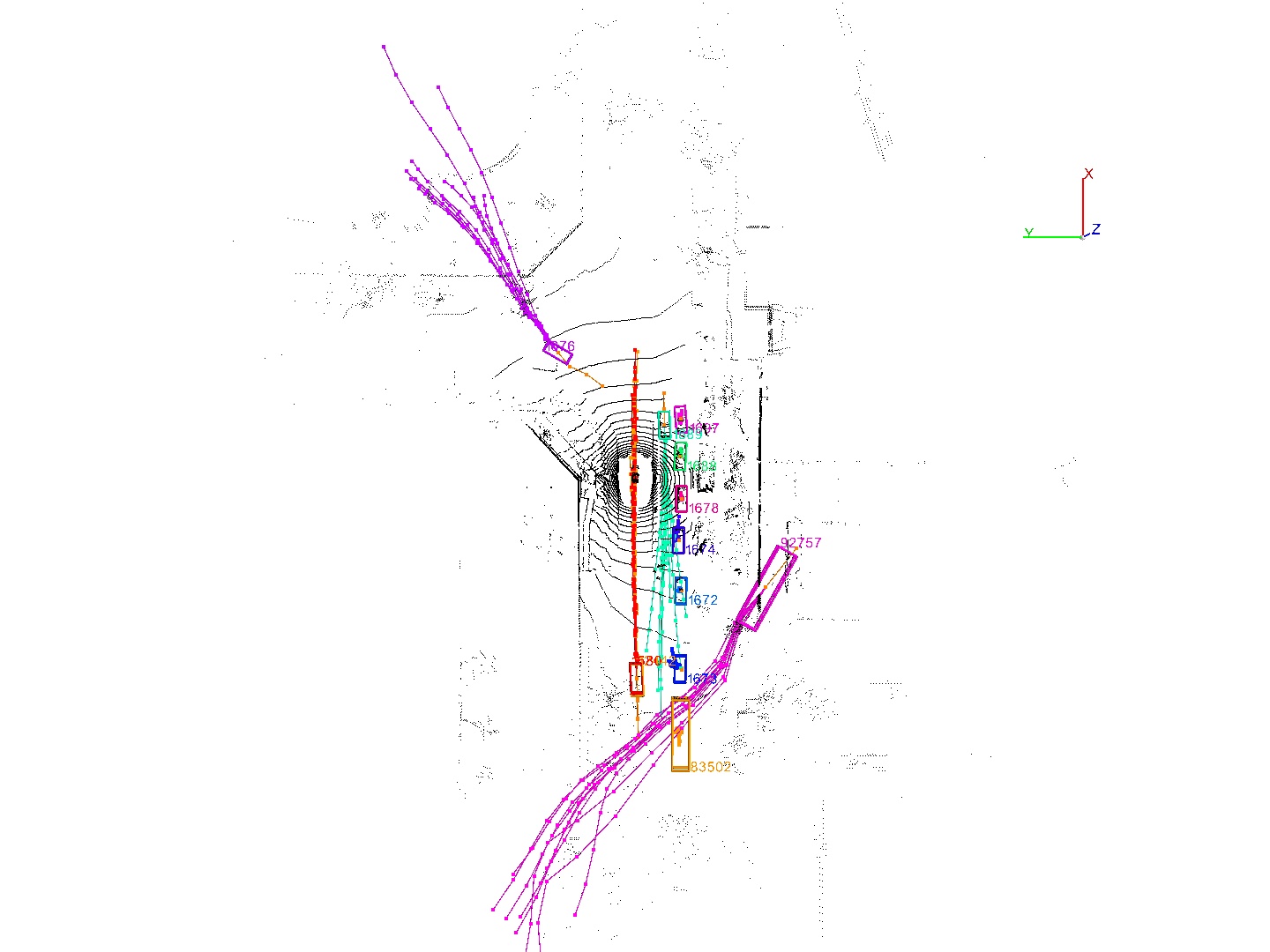}
\vspace{-0.75cm}
\caption{Wrongly-tracked FRAG causes prediction errors. \textbf{(Top)}: When using GT past trajectories as inputs, predictions for the \textcolor{citrine}{yellow} object are accurate. \textbf{(Bottom)}: We show predictions for the same object on the same two frames, but now using past tracklets as inputs. As the detected box of the \textcolor{Purple}{purple} object on the right figure is off by more than 2m from the GT, it causes a wrongly-tracked FRAG error that thwarts predictions.}
\label{fig:FRAG1_prediction}
\vspace{-0.85cm}
\end{center}
\end{figure}

\begin{figure}[t]
\begin{center}
\includegraphics[trim=21cm 32cm 20cm 2.5cm, clip=true, width=0.49\linewidth]{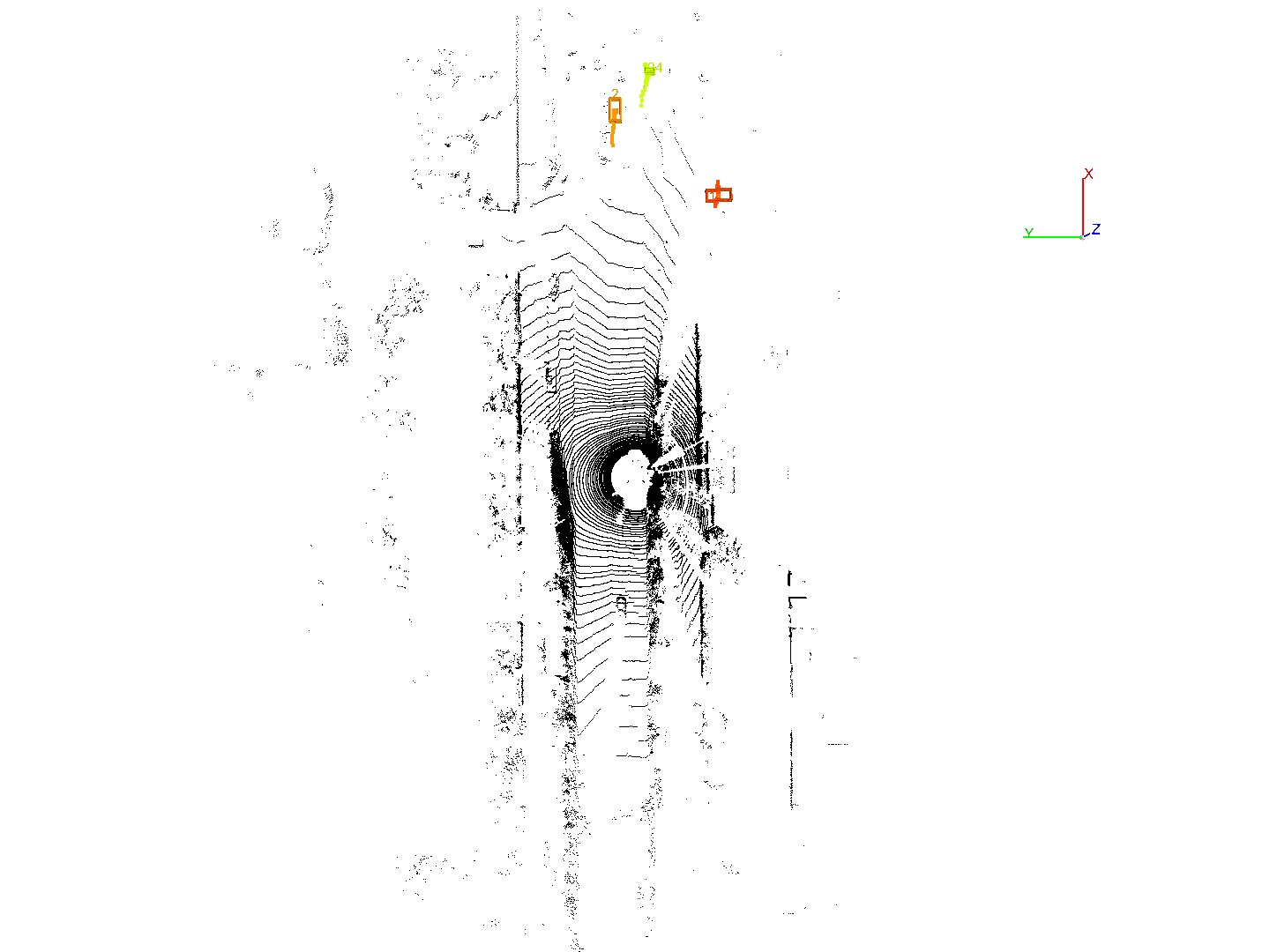}
\includegraphics[trim=21cm 32cm 20cm 2.5cm, clip=true, width=0.49\linewidth]{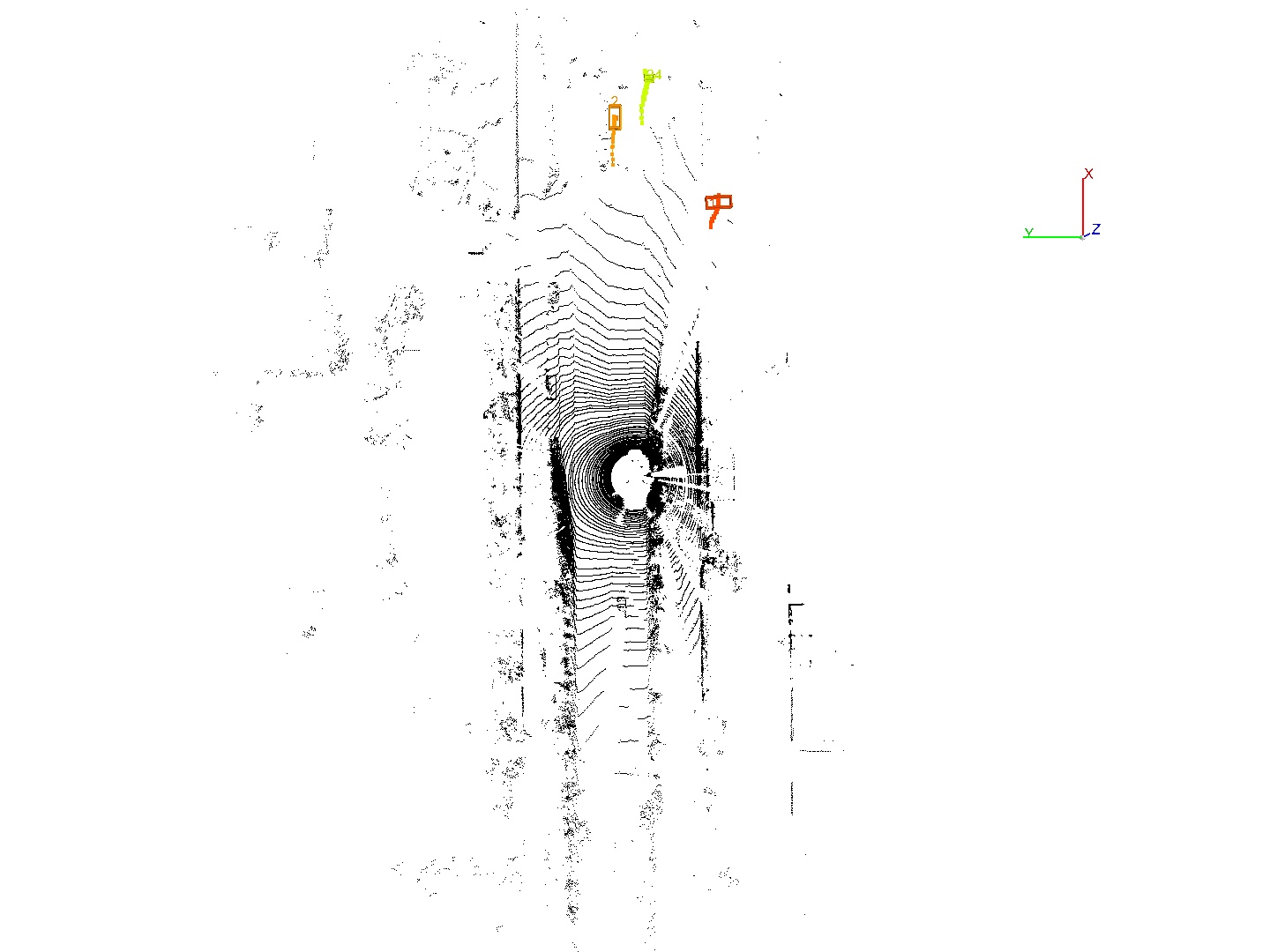}
\includegraphics[trim=21cm 32cm 20cm 2cm, clip=true, width=0.49\linewidth]{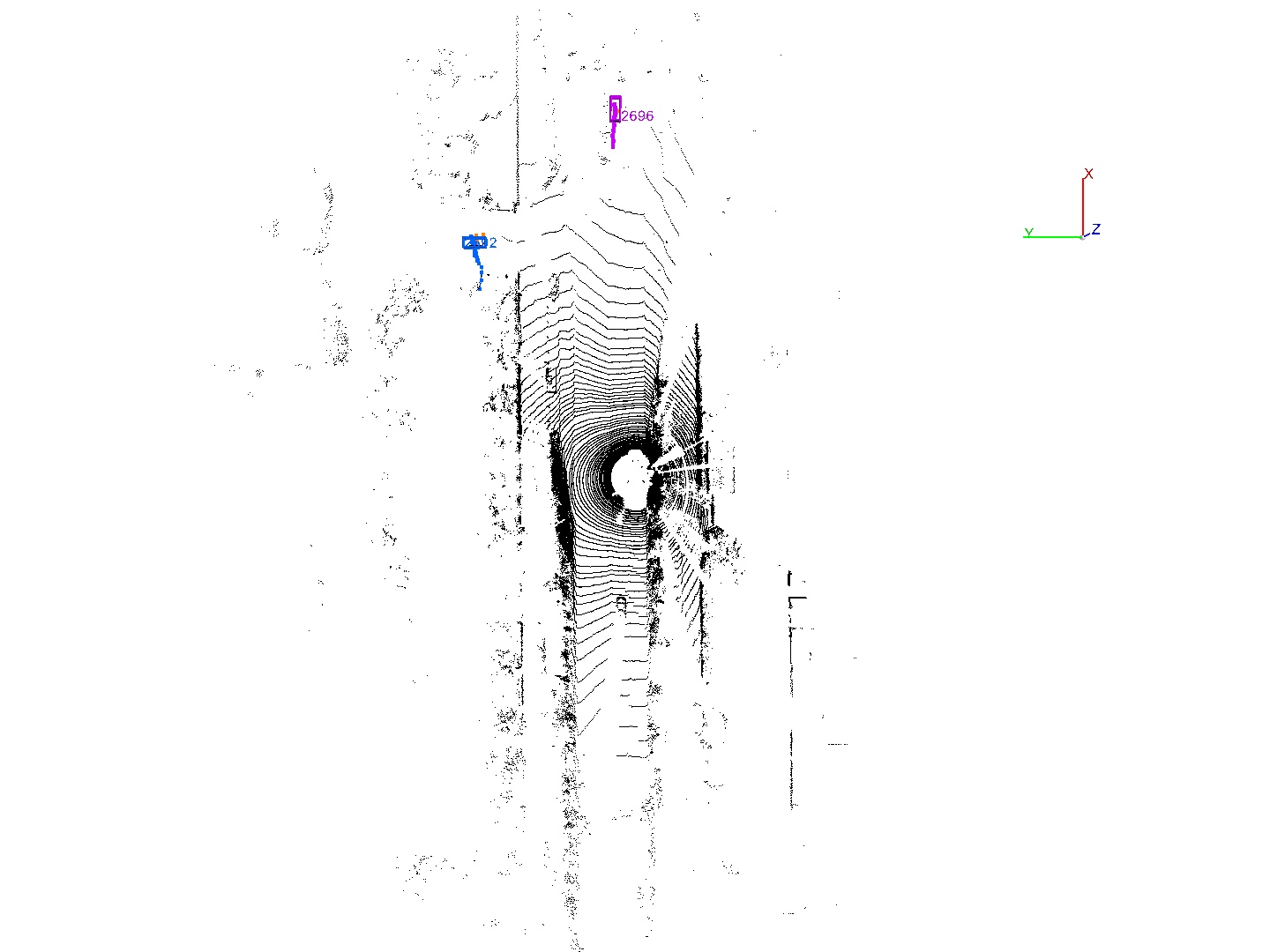}
\includegraphics[trim=21cm 32cm 20cm 2cm, clip=true, width=0.49\linewidth]{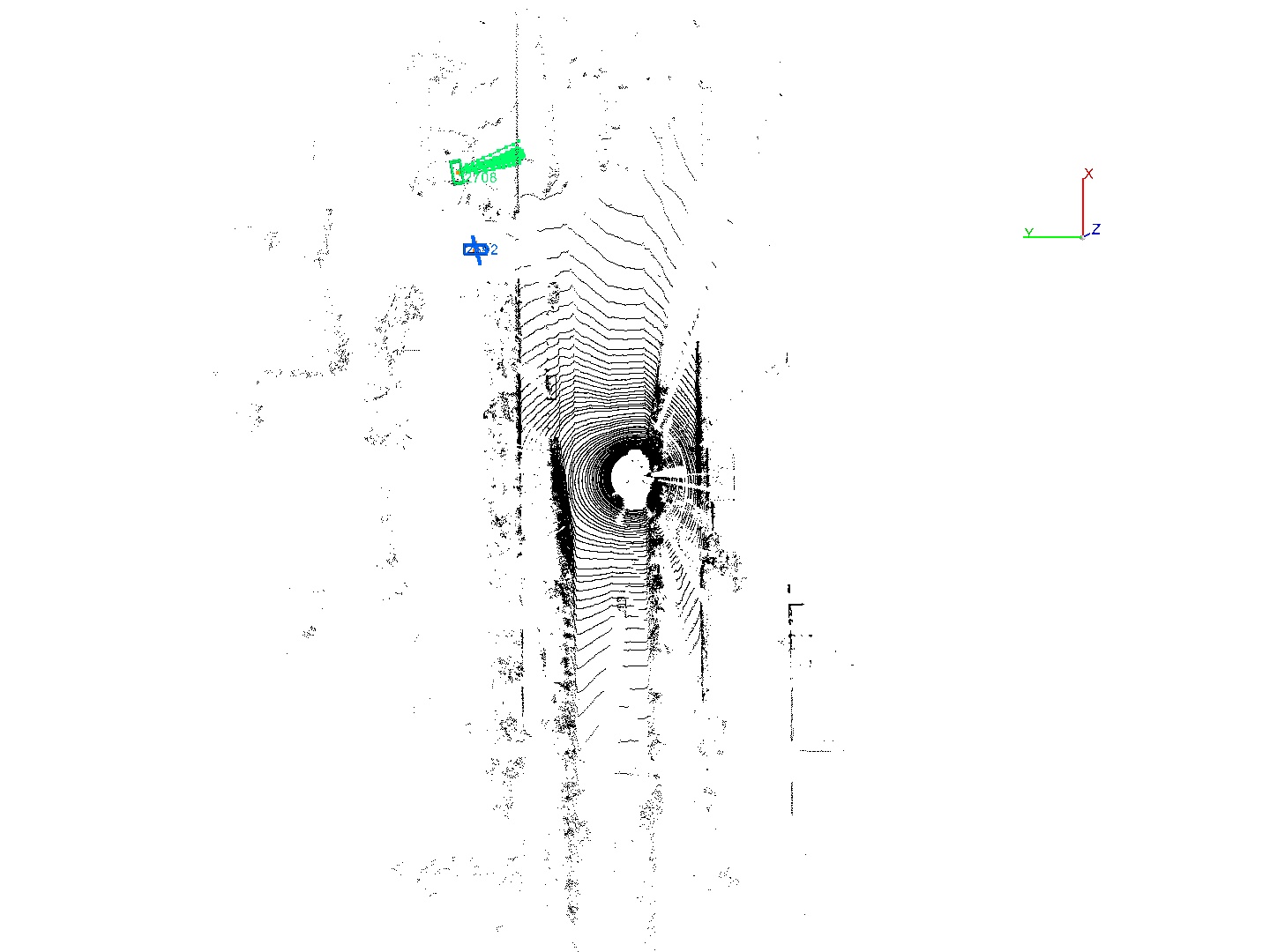}
\vspace{-0.85cm}
\caption{Under-tracked FRAG causes missing predictions. \textbf{(Top)}: We show predictions on two frames when using GT past trajectories as inputs. As GT past trajectories are accurate and stable, predictions are accurate. \textbf{(Bottom)}: We show predictions for the same two objects but now using past tracklets as inputs. As the two objects are missing in the past tracklets (one object is missed in the 2nd frame and the other object is missed in two frames), predictions for these two objects are also missing.}
\label{fig:FRAG2_prediction}
\vspace{-0.8cm}
\end{center}
\end{figure}

\vspace{-0.1cm}
\subsection{Qualitative assessment\label{subsec:qa}}

Leveraging the methodology outlined in Section \ref{sec:eval}, we provide some qualitative insights on the impact of tracking errors on prediction performance.

\vspace{1mm}\noindent\textbf{IDS causes large prediction errors.} As shown in Figure \ref{fig:IDS_prediction} (top), as long as we use GT past trajectories as inputs, predictions are accurate. However, when using past tracklets as inputs, particularly when there is an IDS as shown in Figure \ref{fig:IDS_prediction} (bottom), predictions are off due to the sudden (and erroneous) velocity estimation change in the past tracklet. 

\vspace{1mm}\noindent\textbf{Wrongly-tracked FRAG can also cause errors.} Similar to the case of IDS, a wrongly-tracked FRAG also causes prediction errors. When comparing predictions in Figure \ref{fig:FRAG1_prediction} top (with GT past trajectories as inputs) and bottom (with past tracklets as inputs), one can see that, when the object's past tracklet is slightly off, the corresponding predictions are also off due to the orientation change. 

\vspace{1mm}\noindent\textbf{Under-tracked FRAG causes missing predictions.} Different from the above two cases (which lead to inaccurate predictions), an under-tracked FRAG causes {\em missing} predictions, as there is no past tracklet used as inputs to prediction after the FRAG event. As shown in Figure \ref{fig:FRAG2_prediction}, predictions can be missed for objects with under-tracked FRAG errors.

\vspace{1mm}\noindent\textbf{Spurious tracks cause false positives.} In contrast to under-tracked FRAG, spurious tracks cause predictions that are not supposed to exist, that is, predictions for ghost objects.

\vspace{1mm}\noindent\textbf{Conclusion.} We can see that all tracking errors (IDS, FRAG, and spurious tracks) can lead to prediction errors. In particular, IDS, wrongly-tracked FRAG, and spurious tracks can reduce the precision of the predictions, while under-tracked FRAG and wrongly-tracked FRAG can lead to a lower recall.

\begin{table}[t]
\centering
\vspace{0.25cm}
\caption{Prediction performance for objects with tracking errors.}
\vspace{-0.3cm}
\resizebox{0.485\textwidth}{!}{
\begin{tabular}{@{}lll|rr@{}}
\toprule
Datasets & Eval. Targets (\# of obj) & Inputs to Prediction & $\text{minADE}_{k}$ & $\text{minFDE}_{k}$ \\
\midrule
KITTI       & Objects with IDS (33)         & GT past trajectories      & 0.100 & 0.171\\
            & Objects with IDS (33)         & past tracklets            & 2.820 & 4.514\\
            & Objects with FRAG (330)       & GT past trajectories      & 0.177 & 0.306\\
            & Objects with FRAG (330)       & past tracklets            & 1.621 & 2.155\\
\midrule
nuScenes    & Objects with IDS (4160)       & GT past trajectories      & 0.473 & 0.825\\
            & Objects with IDS (4160)       & past tracklets            & 8.345 & 13.892\\
            & Objects with FRAG (3365)      & GT past trajectories      & 0.621 & 1.108\\
            & Objects with FRAG (3365)      & past tracklets            & 14.520 & 21.815\\
\bottomrule
\label{tab:error_pred}
\end{tabular}}
\vspace{-0.7cm}
\end{table}

\vspace{-0.1cm}
\subsection{Quantitative assessment\label{sec:science}}

In this section, we quantify the impact of IDS/FRAG errors in terms of minADE$_k$ and minFDE$_k$ (we do not consider spurious tracks, as in such cases there is no corresponding GT that can be used to compute ADE/FDE). In particular, for those instances containing IDS/FRAG errors, we compare prediction performance stemming from using GT past trajectories as inputs with prediction performance stemming from using past tracklets as inputs. The results are shown in Table \ref{tab:error_pred}. One can observe that when we replace GT past trajectories with past tracklets as inputs, there is a significant performance drop. In particular, on KITTI IDS instances, there is a 28$\times$ drop from 0.100 to 2.820 in minADE$_k$, and on nuScenes IDS instances, there is an 18$\times$ drop from 0.473 to 8.345 in minADE$_k$. Similar performance drops are observed on FRAG instances. Such performance drops are in agreement with our qualitative findings in Section \ref{subsec:qa} and make predictions arguably almost useless for these objects.

\begin{figure}[t]
\begin{center}
\includegraphics[trim=0.4cm 0.2cm 0.5cm 0.2cm, clip=true, width=0.496\linewidth]{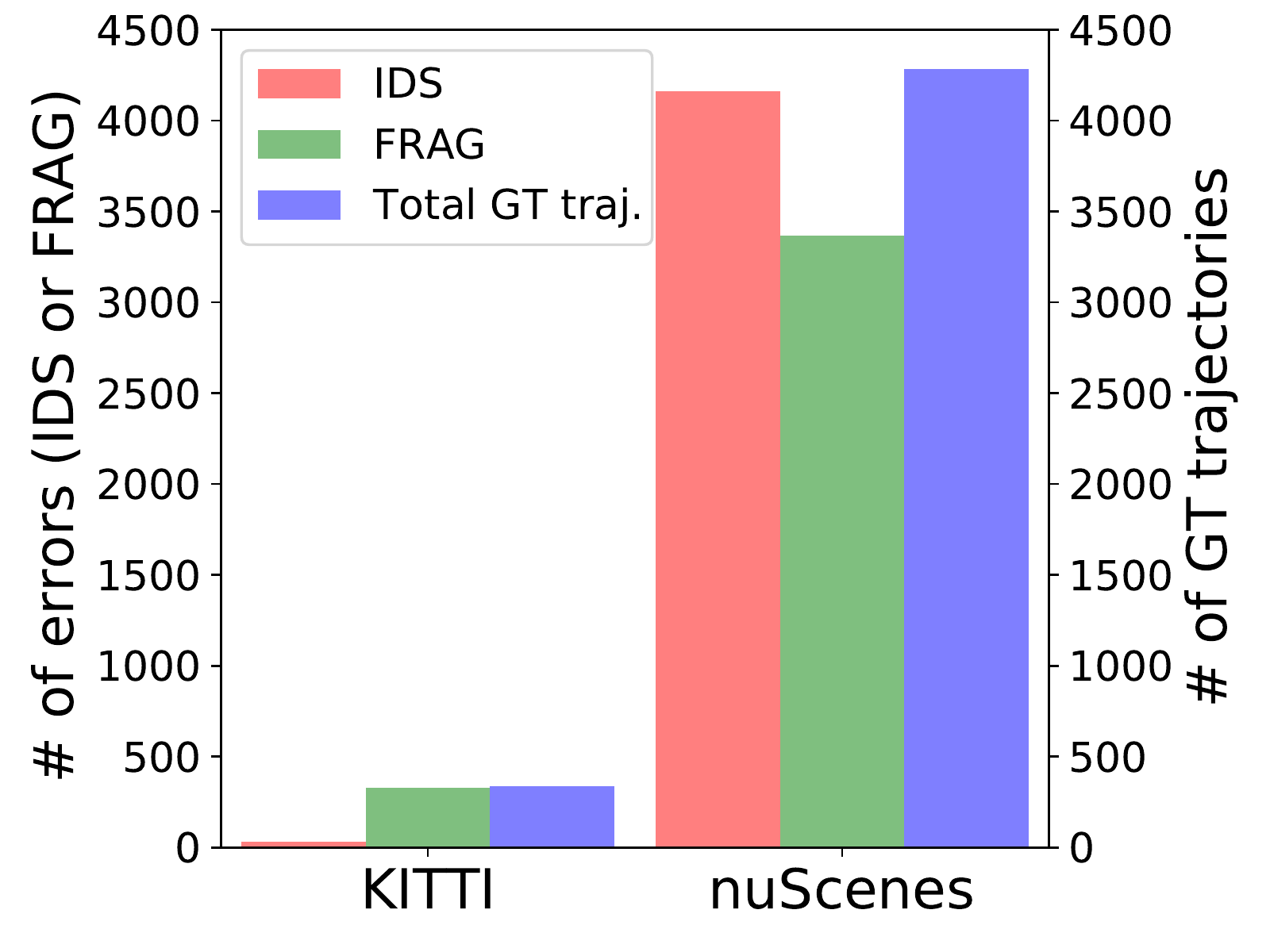}
\includegraphics[trim=1.2cm 1.2cm 0cm 0cm, clip=true, width=0.49\linewidth]{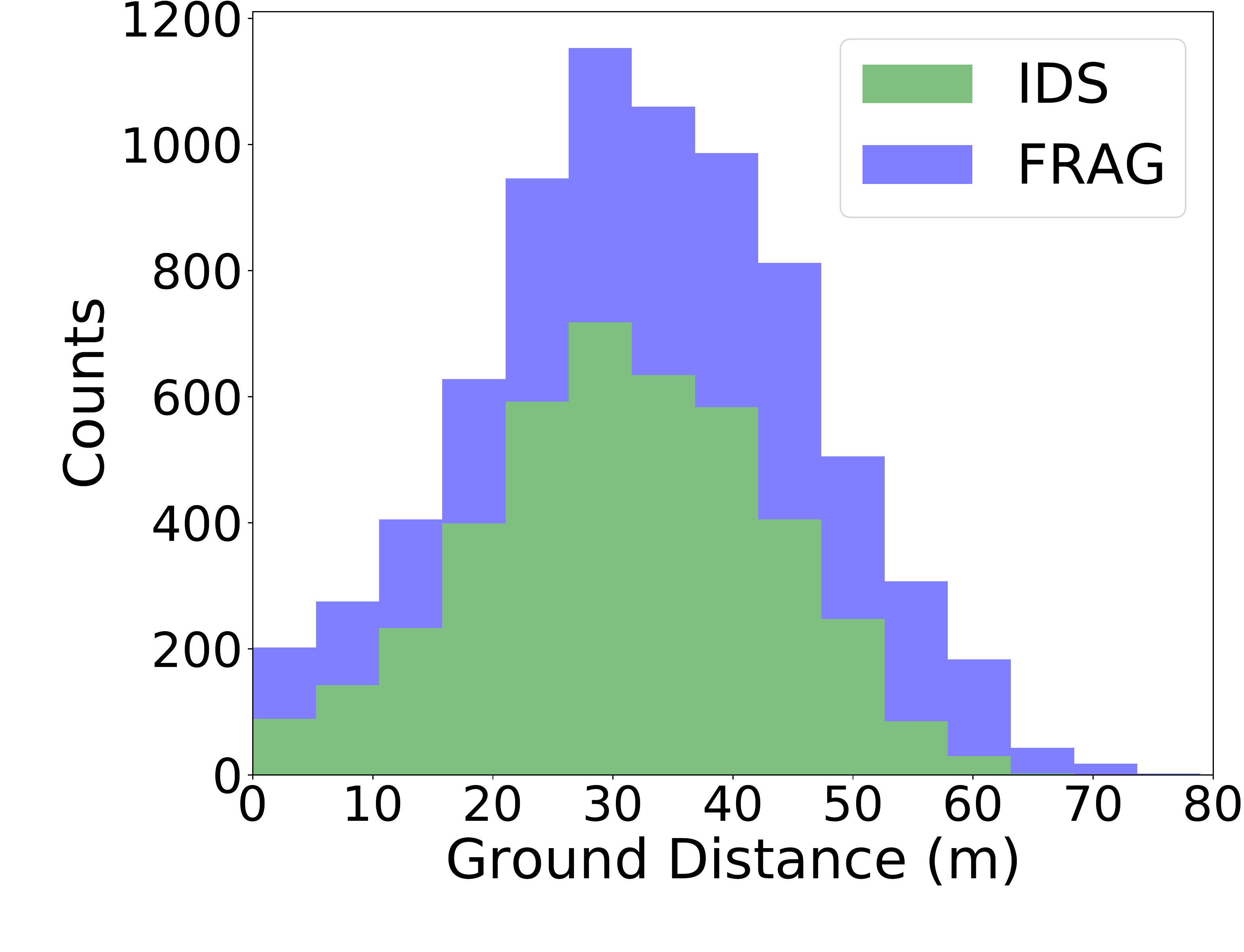}
\vspace{-0.85cm}
\caption{\textbf{(Left)}: IDS/FRAG frequency. On average, every object trajectory may experience a FRAG error on KITTI, and an IDS and/or FRAG error on nuScenes. \textbf{(Right)}: We plot the distribution of distances of erroneously tracked objects with respect to the ego-vehicle on nuScenes. About 200 such objects are close to the ego vehicle (within 5m), that is, at a distance where planning is typically very sensitive to.}
\label{fig:freq}
\vspace{-0.5cm}
\end{center}
\end{figure}

\vspace{-0.1cm}
\subsection{Frequency and spatial distribution assessment}

Sections \ref{subsec:qa} and \ref{sec:science} characterize and quantify how tracking errors can negatively impact prediction performance. But how often do tracking errors happen? And how far are the objects affected by tracking errors from the ego vehicle (that is, do tracking errors also happen for objects very close to the ego vehicle, for which the planning module would be very sensitive to)? Accordingly, we characterize the frequency and spatial distribution of tracking errors below:

\vspace{1mm}\noindent\textbf{IDS/FRAG frequency.} IDS/FRAG cases are indeed quite common. As shown in Figure \ref{fig:freq} (left), on average every trajectory in both the KITTI and nuScenes datasets can yield a FRAG, IDS, or both! The frequency of tracking errors, coupled with their negative impact on prediction (Sections \ref{subsec:qa} and \ref{sec:science}), provide a strong motivation towards developing systematic approaches to account for tracking errors for the purposes of robust prediction (and planning).  

\vspace{1mm}\noindent\textbf{IDS/FRAG spatial distribution.} Planning, in general, is most sensitive to nearby objects. To understand at a conceptual level whether tracking errors can induce erroneous predictions that in turn can thwart planning, we compute the distance of objects experiencing erroneous tracklets from the ego-vehicle. The results are reported in the histogram in Figure \ref{fig:freq} (right). One can see that there is a non-negligible number of IDS/FRAG instances where the tracked object is very close to the ego vehicle. In particular, there are about 1000 IDS/FRAG instances for objects within 15 meters, 500 IDS/FRAG instances for objects within 10 meters, and 200 IDS/FRAG instances for objects within 5 meters. Note that these errors are computed on the nuScenes prediction test set, which only contains 0.83 hours of driving\footnote{The nuScenes prediction test set has 150 sequences with 40 frames per sequence and an FPS of 2, for a total of 3000 seconds $=$ 0.83 hours.}. Thus, we argue that tracking errors can severely hinder safe planning (future research will assess this statement more formally, for example by using planning-aware prediction metrics \cite{Ivanovic2021}). 

\section{Multi-Hypothesis Tracking and Prediction}\label{sec:approach}

\begin{figure}[t]
\begin{center}
\includegraphics[trim=0.3cm 8.5cm 11.5cm 0cm, clip=true, width=\linewidth]{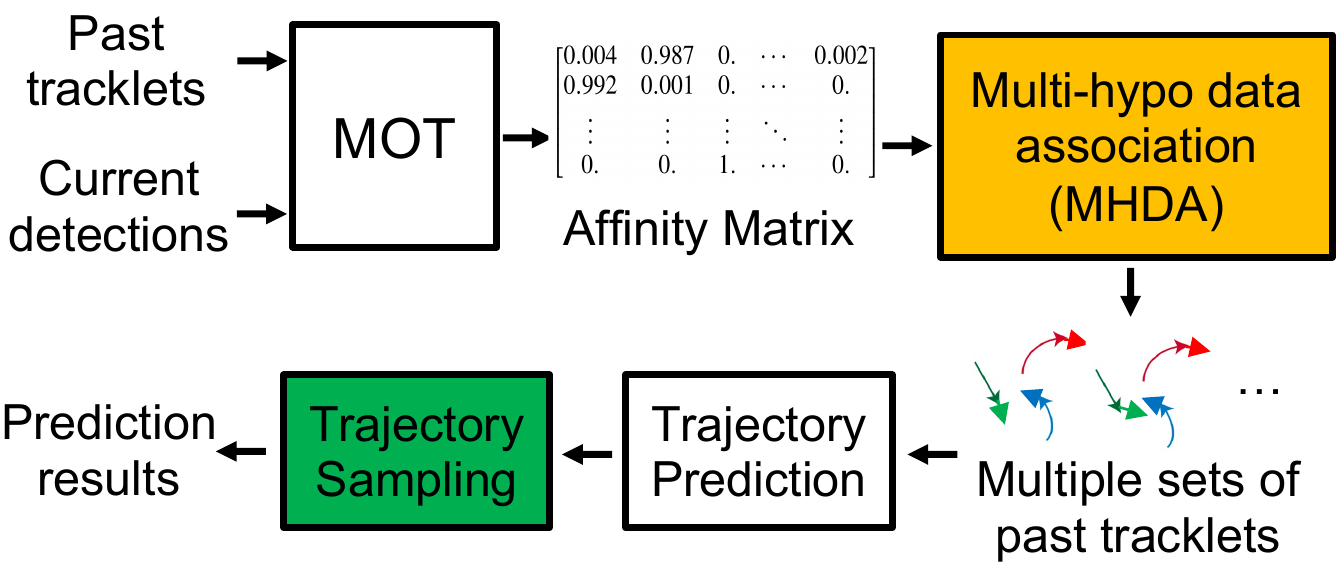}
\vspace{-0.8cm}
\caption{Proposed \methodname framework. The two key modules introduced in \methodname are highlighted in \textcolor{Green}{green} and \textcolor{orange}{orange}. We feed the affinity matrix to MHDA to obtain multiple sets of tracklets. Once predictions are performed on all sets of tracklets, we sample a subset of predictions as final results.}
\label{fig:main}
\vspace{-0.85cm}
\end{center}
\end{figure}

To account for the impact of tracking errors on prediction performance, we propose the \methodname framework which is visualized in Figure \ref{fig:main}. The tracking-prediction pipeline in \methodname is relatively standard, in terms of its modularity and sequence of operations (namely, MOT followed by a prediction module). The two key modules introduced are the MHDA and trajectory sampling, as described below:

\vspace{1mm}\noindent\textbf{Multi-Hypothesis Data Association (MHDA).} The key idea is to reason about multiple hypotheses simultaneously, with the goal of increasing the likelihood of including accurate tracklets that can be used as inputs to downstream prediction. That is, instead of relying on a hard assignment via the Hungarian algorithm, we use MHDA to enlarge the search space and generate sets of plausible tracklets. Explicitly, we use the Murty's H-best assignment \cite{Cox1996}, which maintains $H\geq1$ sets of tracking results at every frame, where each set is referred to as a hypothesis. Typically, the 1st hypothesis is obtained using the Hungarian algorithm by considering the lowest cost, which results in a list of matches between detections and trajectories. To obtain other hypotheses, we tweak the list of matches in the 1st hypothesis by toggling one match at a time in and out of the list, which results in slightly higher costs. After sorting other hypotheses based on costs, the 2nd hypothesis has the 2nd lowest cost and so on. In the case that the 1st hypothesis is erroneous, other hypotheses with slightly higher costs may correspond to a correct association -- thus, by reasoning about multiple hypotheses, the likelihood of retaining accurate tracking results is increased. Each hypothesis (a set of tracklets) is then fed into the prediction module as per Figure \ref{fig:main}.

\vspace{1mm}\noindent\textbf{Trajectory Sampling.} Once we obtain predictions by using each hypothesis as input, we sample a subset of the full set of predictions by computing cluster centers using K-Means++ \cite{Arthur2007}, resulting in a diverse set of predictions. The trajectory sampling step is optional, but useful to limit an excessive number of prediction samples, and helps us with carrying out a fair comparison with single-hypothesis prediction methods. For example, if we use $H=10$ and for each hypothesis we generate $k=20$ prediction samples, there will be 200 samples for each object. In this case, we would sample only $20$ samples out of the 200 to carry out a fair comparison with single-hypothesis prediction methods using $k=20$ samples. 

\vspace{1mm}\noindent\textbf{Conclusion.} In summary, our \methodname framework can be applied to any tracking-prediction pipeline that is based on single-hypothesis matching -- the main modification is to replace the matching algorithm with MHDA. 

\section{Experiments}

As \methodname is designed to improve prediction, we follow the standard prediction evaluation as described in Section \ref{sec:eval}. For additional implementation details and hyper-parameters, we refer the reader to our code. Here, we categorize our prediction experiments into: 1) targeted evaluation, which analyzes prediction performance for objects affected by tracking errors, and 2) global evaluation, which analyzes prediction performance across all tracked objects, whether or not they are affected by tracking errors. Also, we provide a runtime speed analysis. The key takeaway is that \methodname improves both targeted and global prediction performance, with a relatively minor computation overhead.

\vspace{1mm}\noindent\textbf{Targeted evaluation.} Results are provided in Table \ref{tab:quan_ids} in terms of minADE$_k$ and minFDE$_k$. The first row of each block corresponds to the standard single-hypothesis tracking-prediction pipeline (AB3DMOT+PTP), which we refer to as STP. As shown in Section \ref{sec:science}, STP yields large prediction errors due to IDS/FRAG. Next, we see that \methodname significantly improves prediction performance on both the KITTI and nuScenes datasets. Specifically, when using $k=400$ prediction samples from all $H=20$ hypotheses, we see a 4$\times$ minADE$_{k}$ performance boost on KITTI IDS (\emph{i.e.}, from 2.820 to 0.707), a 19.5\% minADE$_{k}$ performance boost on KITTI FRAG, a 2.5$\times$ performance boost on nuScenes IDS, and a 2$\times$ performance boost on nuScenes FRAG. To compare \methodname and STP under the same number of samples $k$ (and thus avoid giving \methodname an unfair advantage with a larger number of samples), we apply trajectory sampling to \methodnamenosp. Remarkably, one can see that prediction performance after sampling is only slightly lower (\emph{e.g.}, minADE$_k$ raises from 0.707 to 0.747 on KITTI IDS), meaning that the proposed trajectory sampling scheme generally retains accurate tracklets. Also, as more hypotheses are used, better performance is achieved (compare the cases with $H=5$, $10$, and $20$). Importantly, even when using only 5 hypotheses, prediction performance is improved by 2.5$\times$ on KITTI IDS cases. 

Finally, even though the minADE$_k$ and minFDE$_k$ metrics are not suitable to characterize how \methodname improves prediction performance on spurious track instances, it is easy to argue why this is the case. Indeed, the likelihood of removing spurious tracks is increased under \methodnamenosp, as different hypotheses have different matching results, and some hypotheses may not associate false positive detections to trajectories.

\vspace{1mm}\noindent\textbf{Global evaluation.} Results are provided in Table \ref{tab:quan_all}. Again, \methodname largely improves performance over STP, \emph{e.g.}, minFDE$_k$ from 0.278 to 0.238 (14.4\% improvement) on KITTI, and minFDE$_k$ from 3.819 to 2.512 (34.2\%) on nuScenes. Improvement on nuScenes is larger as there is a higher percentage of IDS/FRAG instances. In brief, \methodname significantly improves both targeted {\em and} global prediction performance. 

Note that although we follow the nuScenes evaluation protocol, the ADE/FDE numbers in Table \ref{tab:quan_all} are not comparable to the numbers on the nuScenes leaderboard as we consider past tracklets as inputs (as opposed to GT past trajectories, as is the case for the nuScenes leaderboard results).

\begin{table}[t]
\vspace{0.2cm}
\centering
\caption{Prediction performance for objects with IDS/FRAG.}
\vspace{-0.3cm}
\resizebox{0.49\textwidth}{!}{
\begin{tabular}{@{}lll|rr@{}}
\toprule
Datasets & Targets & Methods & $\text{minADE}_{k}$ & $\text{minFDE}_{k}$\\
\midrule
KITTI       & IDS   & STP, H=1, k=20                                & 2.820 & 4.514 \\
\cmidrule{3-5}
            &       & \methodname (Ours), H=5, k=100                & 1.099 & 1.768 \\
            &       & \methodname (Ours), H=10, k=200               & 0.844 & 1.332 \\
            &       & \methodname (Ours), H=20, k=400               & 0.707 & 1.093 \\
            &       & \methodname (Ours), H=5, k=20, sampling       & 1.118 & 1.802 \\
            &       & \methodname (Ours), H=10, k=20, sampling      & 0.876 & 1.390 \\
            &       & \methodname (Ours), H=20, k=20, sampling      & 0.747 & 1.173 \\
\midrule            
KITTI       & FRAG  & STP, H=1, k=20                                & 1.621 & 2.155 \\
\cmidrule{3-5}
            &       & \methodname (Ours), H=5, k=100                & 1.436 & 1.862 \\
            &       & \methodname (Ours), H=10, k=200               & 1.385 & 1.765 \\
            &       & \methodname (Ours), H=20, k=400               & 1.305 & 1.627 \\
            &       & \methodname (Ours), H=5, k=20, sampling       & 1.448 & 1.888 \\
            &       & \methodname (Ours), H=10, k=20, sampling      & 1.404 & 1.801 \\
            &       & \methodname (Ours), H=20, k=20, sampling      & 1.335 & 1.688 \\
\midrule
nuScenes    & IDS   & STP, H=1, k=10                                & 8.345 & 13.892 \\
\cmidrule{3-5}
            &       & \methodname (Ours), H=10, k=100               & 4.143 &  6.464 \\
            &       & \methodname (Ours), H=20, k=200               & 3.321 &  5.052 \\
            &       & \methodname (Ours), H=10, k=10, sampling      & 4.573 &  7.303 \\
            &       & \methodname (Ours), H=20, k=10, sampling      & 3.923 &  6.210 \\
\midrule
nuScenes    & FRAG  & STP, H=1, k=10                                & 14.520 & 21.815 \\
\cmidrule{3-5}
            &       & \methodname (Ours), H=10, k=100               &  9.017 & 12.721 \\
            &       & \methodname (Ours), H=20, k=200               &  7.697 & 10.606 \\
            &       & \methodname (Ours), H=10, k=10, sampling      &  9.585 & 13.846 \\
            &       & \methodname (Ours), H=20, k=10, sampling      &  8.476 & 12.105 \\
\bottomrule
\label{tab:quan_ids}
\end{tabular}}
\vspace{-1.1cm}
\end{table}
            
\begin{figure}[t]
\begin{center}
\includegraphics[trim=1.6cm 1.3cm 0cm 0cm, clip=true, width=0.49\linewidth]{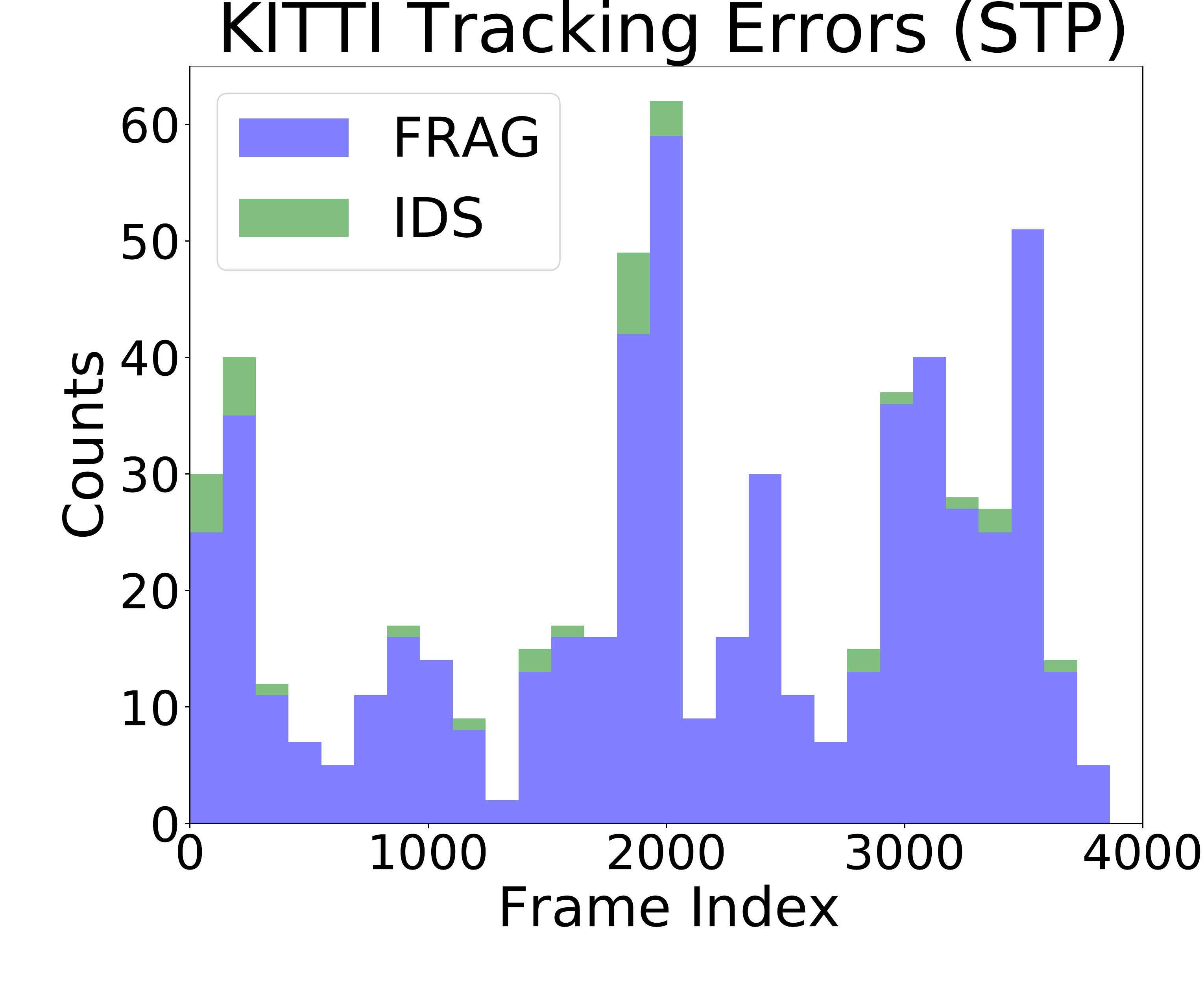}
\includegraphics[trim=1.6cm 1.3cm 0cm 0cm, clip=true, width=0.49\linewidth]{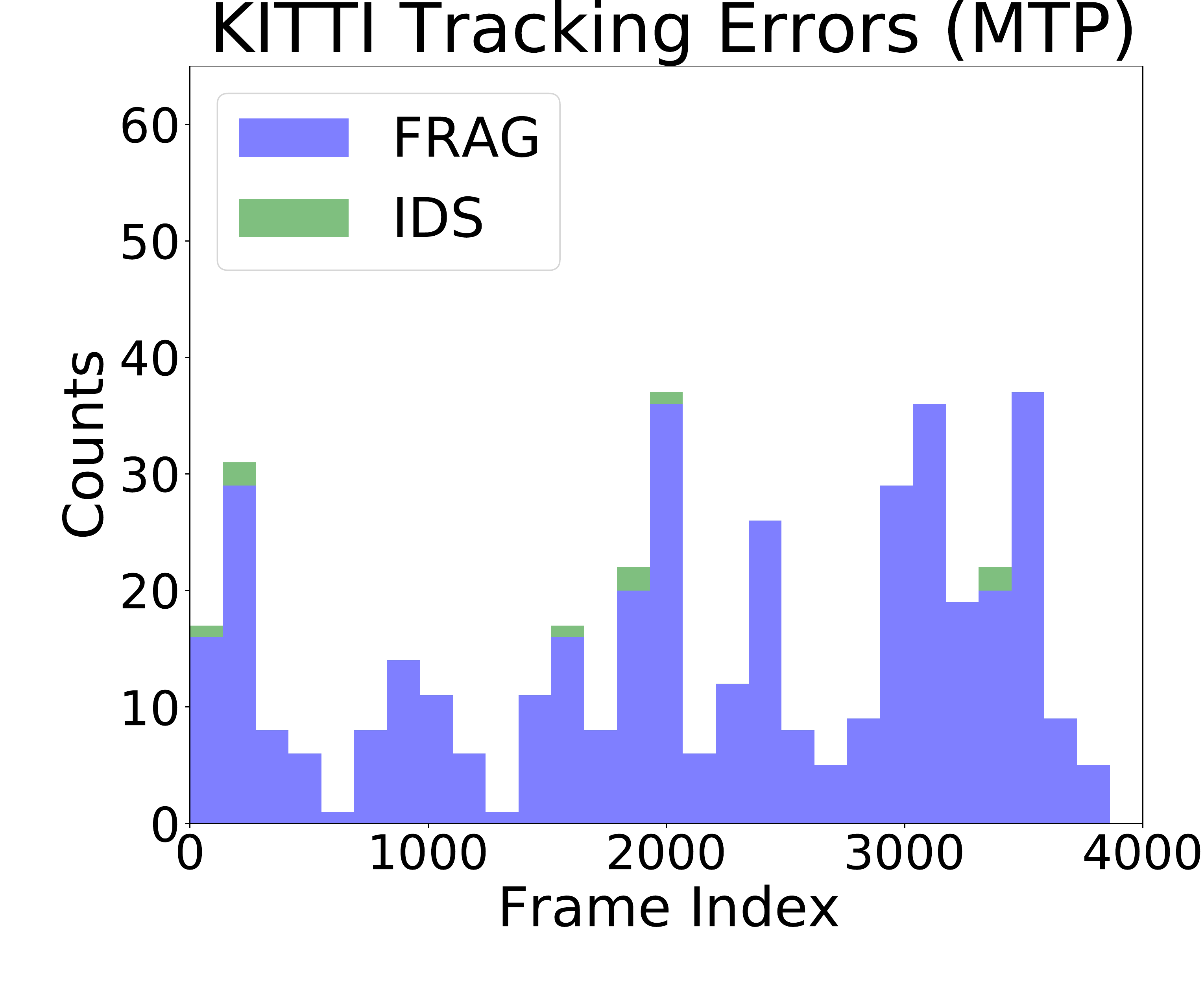}\\
\includegraphics[trim=1.6cm 1.3cm 0cm 0cm, clip=true, width=0.49\linewidth]{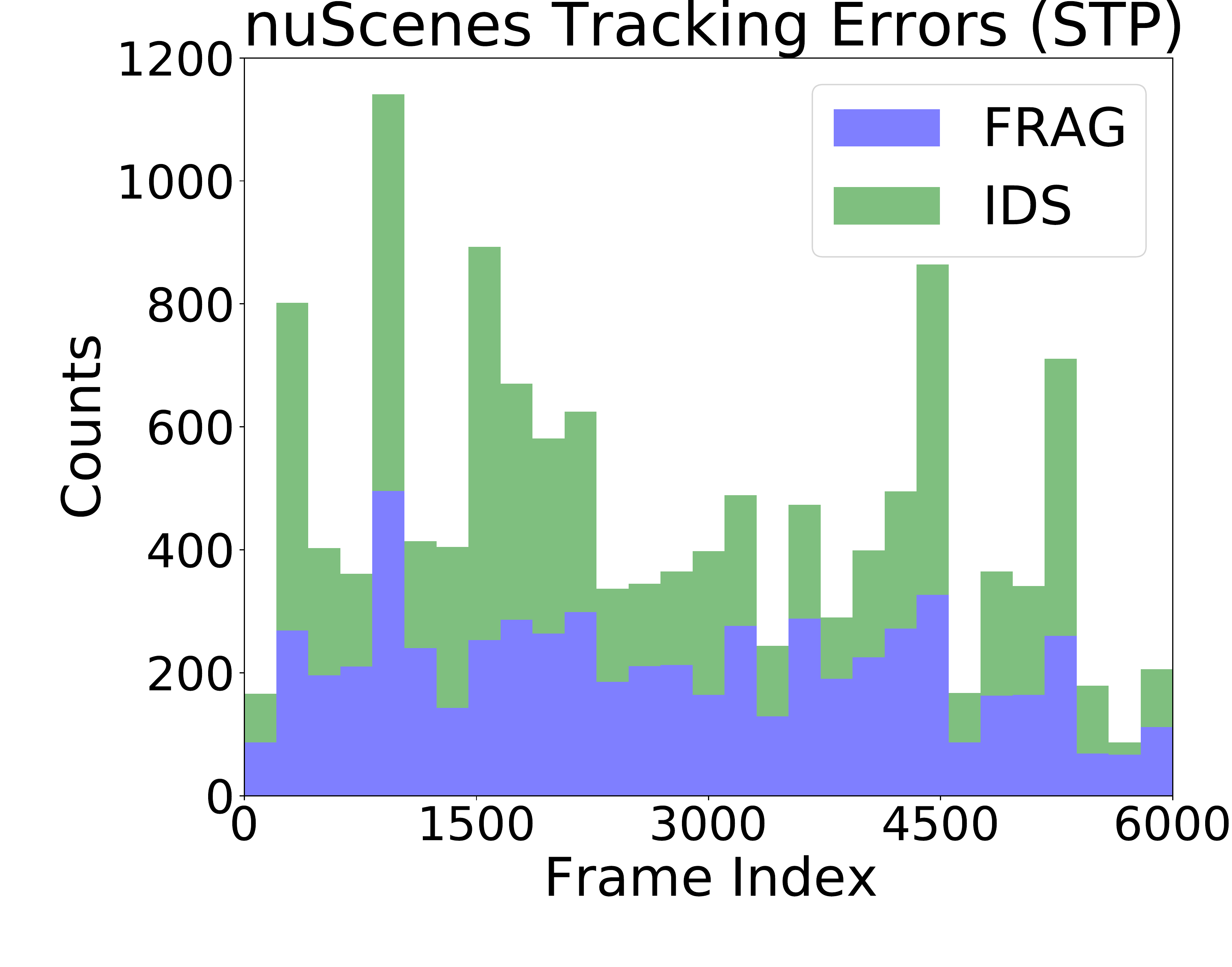}
\includegraphics[trim=1.6cm 1.3cm 0cm 0cm, clip=true, width=0.49\linewidth]{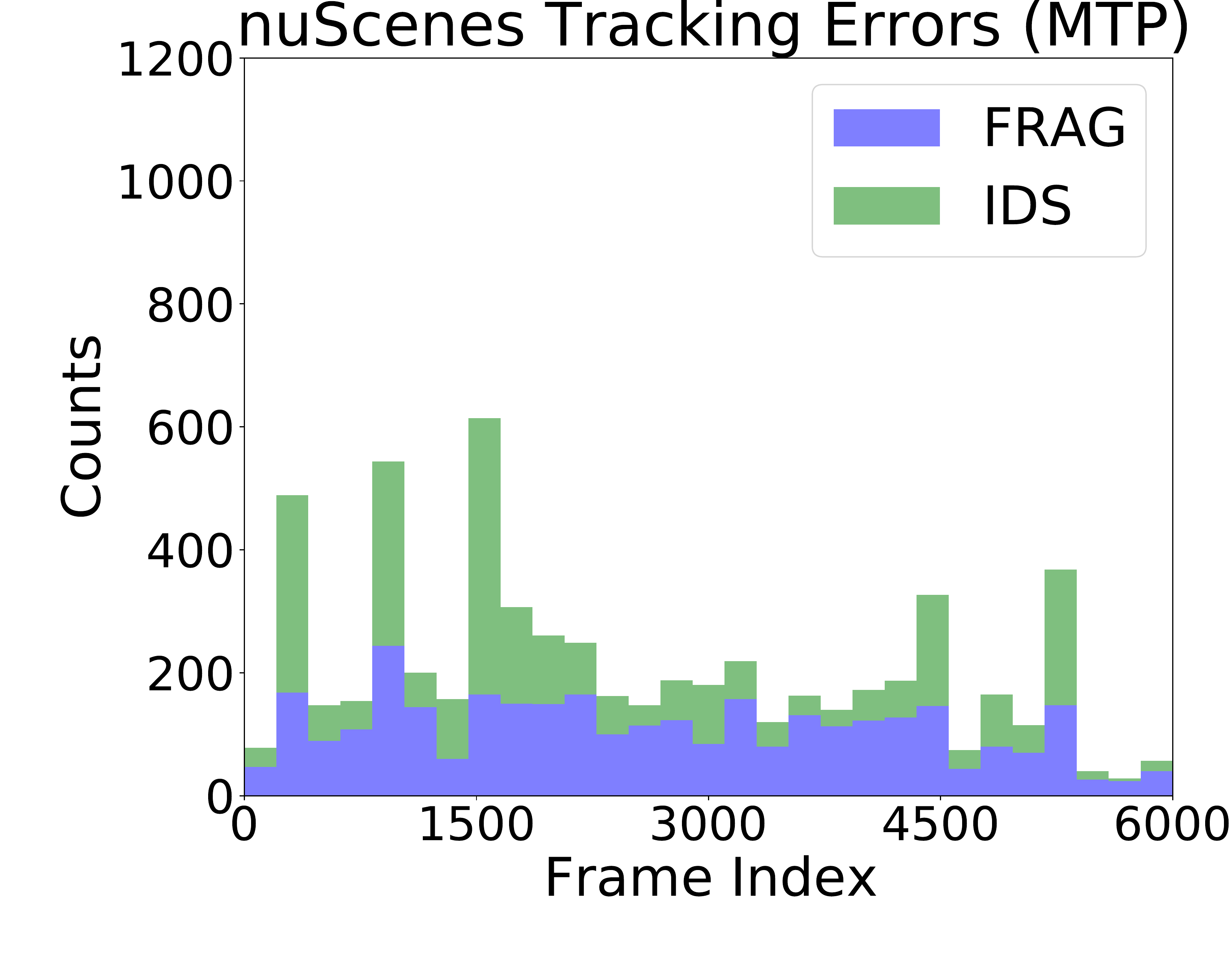}
\vspace{-0.85cm}
\caption{We plot the distribution of IDS/FRAG errors over frames for the STP (left) and \methodname (right). It is clear that tracking errors are largely reduced when considering all 20 hypotheses in \methodname on both KITTI and nuScenes.}
\label{fig:error_sta}
\vspace{-0.6cm}
\end{center}
\end{figure}

\vspace{1mm}\noindent\textbf{Tracking Error Statistics.} To gain insights on why \methodname improves prediction performance, we show an intermediate tracking error analysis in Figure \ref{fig:error_sta} for IDS/FRAG instances on nuScenes and KITTI. Specifically, we plot the distribution over frames of IDS/FRAG instances for STP on the left, and the distribution over frames of IDS/FRAG instances present in {\em all} of the $H=20$ hypotheses for \methodname on the right. One can notice that a large portion of tracking errors in STP does not exist in at least one of the hypotheses being considered by \methodnamenosp. If we count the number of FRAG/IDS over all frames, the 33 IDS instances experienced by STP on KITTI are reduced to only 9 that are shared by all 20 hypotheses under \methodname (72.7\% reduction), and the 7083 IDS instances experienced by STP on nuScenes are reduced to only 2835 that are shared by all 20 hypotheses under \methodname (60.0\% reduction). FRAG errors are reduced by a similar amount. This provides strong justification for the inclusion of MHDA in a tracking-prediction pipeline.   

\begin{table}[t]
\centering
\vspace{0.2cm}
\caption{Prediction performance for all objects being predicted.}
\vspace{-0.3cm}
\resizebox{0.49\textwidth}{!}{
\begin{tabular}{@{}ll|rr@{}}
\toprule
Datasets & Methods & $\text{minADE}_{k}$ & $\text{minFDE}_{k}$ \\
\midrule
KITTI       & STP, H=1, k=20                            & 0.185 & 0.278 \\
\cmidrule{2-4}
            & \methodname (Ours), H=5, k=100            & 0.163 & 0.235 \\
            & \methodname (Ours), H=10, k=200           & 0.152 & 0.215 \\
            & \methodname (Ours), H=20, k=400           & 0.146 & 0.203 \\
            & \methodname (Ours), H=5, k=20, sampling   & 0.170 & 0.252 \\
            & \methodname (Ours), H=10, k=20, sampling  & 0.164 & 0.240 \\
            & \methodname (Ours), H=20, k=20, sampling  & 0.162 & 0.238 \\
\midrule
nuScenes    & STP, H=1, k=10                            & 2.320 & 3.819 \\
\cmidrule{2-4}
            & \methodname (Ours), H=10, k=100           & 1.498 & 2.293 \\
            & \methodname (Ours), H=20, k=200           & 1.325 & 1.979 \\
            & \methodname (Ours), H=10, k=10, sampling  & 1.691 & 2.692 \\
            & \methodname (Ours), H=20, k=10, sampling  & 1.585 & 2.512 \\
\bottomrule
\label{tab:quan_all}
\end{tabular}}
\vspace{-0.65cm}
\end{table}

\begin{table}[t]
\vspace{0.2cm}
\centering
\caption{Runtime speed for tracking and prediction on KITTI (FPS).}
\vspace{-0.35cm}
\resizebox{0.49\textwidth}{!}{
\begin{tabular}{@{}lrrrr@{}}
\toprule
& STP (H=1) & \methodname (H=5) & \methodname (H=10) & \methodname (H=20) \\
\midrule
3D MOT      & 207.4 & 65.2 & 24.6 & 8.1 \\
\midrule
Prediction  & 6.5 & 6.5 & 6.3 & 5.8 \\
\bottomrule
\label{tab:speed}
\end{tabular}}
\vspace{-0.3cm}
\end{table}

\vspace{1mm}\noindent\textbf{Runtime Speed.} MHDA unavoidably introduces a computation overhead, which is characterized in terms of FPS on the KITTI dataset in Table \ref{tab:speed}. It is expected that, as $H$ increases, tracking takes longer. The good news is that, even with $H=10$, tracking runtime is still acceptable (near real-time), without requiring a GPU implementation. This can be attributed to the excellent speed of \cite{Weng2020_AB3DMOT}. Interestingly, there is nearly no runtime degradation for prediction as $H$ increases. This is because predictions for different hypotheses are completely independent, so one can easily run them in parallel, although more GPU memory is needed ($\approx$1.8Gb per hypothesis is required by \cite{Weng2021_PTP}). Runtime for K-means++ sampling is negligible and so it is not included in the table.

\begin{table}[t]
\centering
\vspace{-0.3cm}
\caption{Global prediction evaluation on WILDTRACK.}
\vspace{-0.3cm}
\resizebox{0.49\textwidth}{!}{
\begin{tabular}{@{}ll|rr@{}}
\toprule
Methods & Parameters & $\text{minADE}_{k}$ & $\text{minFDE}_{k}$ \\
\midrule
STP, \cite{Bewley2016} + \cite{Goel2016}    & H=1, k=1              & 1.024 & 1.335 \\
\midrule
Re-tracking, \cite{Yu2021}                  & H=1, k=1              & 0.893 & 1.241 \\
\midrule
MTP (Ours)                                  & H=20, k=1, sampling   & 0.852 & 1.086 \\
MTP (Ours) + Re-tracking                    & H=20, k=1, sampling   & 0.770 & 0.967 \\
\bottomrule
\label{tab:wildtrack}
\end{tabular}}
\vspace{-1cm}
\end{table}            
        
\vspace{1mm}\noindent\textbf{Comparison with Concurrent Work \cite{Yu2021}.} As discussed in Section \ref{sec:rw}, an unpublished work \cite{Yu2021} has proposed a single-hypothesis-based re-tracking solution to mitigate the impact of tracking errors on prediction. Though \cite{Yu2021} has open-sourced its code, direct comparison between \methodname and \cite{Yu2021} is not immediate as \cite{Yu2021} is implemented on a 2D (rather than 3D, as in this paper) detection, tracking, and prediction pipeline, that is, using MaskRCNN \cite{He2017} for detection in images, using SORT \cite{Bewley2016} for 2D MOT, and using Social-LSTM \cite{Goel2016} for bird's eye view trajectory prediction. The method is evaluated on a multi-camera dataset, WILDTRACK \cite{Chavdarova2018}. To ensure a fair comparison, we add our MHDA and trajectory sampling modules to the SORT+Social-LSTM pipeline implemented by \cite{Yu2021} and carry out the evaluation on WILDTRACK. Note that, as \cite{Yu2021} did not release detection results by the time this paper was submitted, we use the Detectron2 implementation of MaskRCNN with an X101-FPN backbone \cite{detectron2} to generate detections. Prediction results are shown in Table \ref{tab:wildtrack}, in terms of minADE$_k$ and minFDE$_k$. Here, $k=1$ as Social-LSTM is a deterministic prediction approach. Both the \methodname and re-tracking approaches \cite{Yu2021} show improvement over STP when using past tracklets as inputs, with \methodname showing a slightly larger improvement. Importantly, the re-tracking approach, which is single-hypothesis-based, and the \methodname framework are complementary, and indeed combining the two approaches (in the last row in Table \ref{tab:wildtrack}) further improves prediction performance!

\section{CONCLUSIONS}

In this paper, we studied how tracking errors can impact prediction performance via qualitative and quantitative analyses. These analyses led to the design of the \methodname framework, which simultaneously reasons about multiple sets of tracking results in order to account for tracking errors. We demonstrated how \methodname significantly improves prediction performance, particularly in those instances containing tracking errors -- all for a relatively minor computation overhead. 

This work opens up a number of future research directions. First, it is of interest to better understand how to optimally choose $H$ and $k$ to consider as a function of computational requirements and target operational design domains. Second, it is of interest to extend our analysis by considering additional methods for MOT and prediction, and planning-aware evaluation metrics (quantitatively assessing how tracking errors ultimately impact planning). Third, the \methodname framework is quite general and can be augmented with other techniques aimed at mitigating the propagation of tracking errors. This provides an exciting opportunity to make predictions even more robust to tracking errors. Finally, we plan to study how errors and uncertainty propagate across other modules within the autonomy stack, with the ultimate goal of devising more robust autonomy stacks. 

\addtolength{\textheight}{-1.5cm}   
%%%%%%%%%%%%%%%%%%%%%%%%%%%%%%%%%%%%%%%%%%%%%%%%%%%%%%%%%%%%%%%%%%%%%%%%%%%%%%%%

\bibliographystyle{IEEEtran}
\bibliography{IEEEabrv,main}

\end{document}